\documentclass[preprint,12pt]{elsarticle}
\usepackage{mathtools,amssymb, bm}
\usepackage[ruled,vlined]{algorithm2e}
\usepackage[noend]{algpseudocode}
\usepackage{amsfonts}
\usepackage{natbib}
\usepackage{float}
\usepackage{hyperref}
\usepackage{xcolor}
\usepackage{lineno}
\usepackage{amsthm}
\usepackage{amsmath}
\usepackage{mathrsfs}

\usepackage{bm}
\usepackage{ragged2e}
\justifying
\usepackage{amsmath}
\usepackage{graphicx}
\graphicspath{{{./}}}

\newcommand{\figwidth}{0.65\textwidth}
\newcommand{\etal}{{\it et al.}}
\newcommand{\ie}{{ i.e.}}

\usepackage{xcolor}

\newcommand{\Aref}[1]{Alg.\,~\ref{#1}}
\newcommand{\aref}[1]{\ref{#1}}
\newcommand{\fref}[1]{Fig.\,~\ref{#1}}
\newcommand{\tref}[1]{Table\,~\ref{#1}}
\newcommand{\eref}[1]{Eq.\,~(\ref{#1})}

\newcommand{\sref}[1]{Sec.\!~\ref{#1}}

\newcommand{\cref}[1]{Ref.\,~\cite{#1}}

\newcommand{\aposteriori}{{\it{a posteriori}} }

\newcommand{\bs}{\mathsf{b}}

\newcommand{\zs}{\mathsf{z}}

\newcommand{\Hs}{\mathsf{H}}

\newcommand{\Vs}{\mathsf{V}}
\newcommand{\Ws}{\mathsf{W}}
\newcommand{\Xs}{\mathsf{X}}

\newcommand{\Ps}{\mathsf{P}}

\newcommand{\eb}{\mathbf{e}}

\newcommand{\Cb}{\mathbf{C}}
\newcommand{\Eb}{\mathbf{E}}
\newcommand{\Fb}{\mathbf{F}}

\newcommand{\Sb}{\mathbf{S}}
\newcommand{\Ib}{\mathbf{I}}

\newcommand{\Uc}{\mathcal{U}}
\newcommand{\Nc}{\mathcal{N}}
\newcommand{\Lc}{\mathcal{L}}

\newcommand{\etab}{{\boldsymbol{\eta}}}

\newcommand{\Sigmab}{\boldsymbol{\Sigma}}

\newcommand{\phib}{{\boldsymbol{\phi}}}
\newcommand{\psib}{{\boldsymbol{\psi}}}
\newcommand{\Psib}{\boldsymbol{\Psi}}

\newcommand{\grad}{{\boldsymbol{\nabla}}}
\newcommand{\partialb}{{\boldsymbol{\partial}}}

\newcommand{\tr}{\operatorname{tr}}

\newcommand{\argmin}{\operatorname{argmin}}

\newcommand{\NN}{\mathsf{N}\!\mathsf{N}}

\newcommand{\parameters}{{\boldsymbol{\theta}}}
\newcommand{\data}{\mathcal{D}}
\newcommand{\inputvector}{\mathsf{x}}
\newcommand{\outputvector}{\mathsf{y}}
\newcommand{\hiddenvector}{\mathsf{h}}

\newcommand{\prob}{\pi}

\newcommand{\lik}{L}
\newcommand{\giv}{\, | \,}

\newcommand{\kernel}{\kappa}

\newcommand{\KLdiv}{\operatorname{KL}}

\newcommand{\CDF}{\operatorname{CDF}}
\newcommand{\sig}{\operatorname{sig}}

\newcommand{\loss}{\mathcal{L}}

\newcommand{\defgrad}{\mathbf{F}}
\date{}
\makeatletter

\journal{}
\begin{document}
\begin{frontmatter}
\title{{\bf Improving the performance of Stein variational inference  through extreme sparsification of physically-constrained neural network models}}
\author{Govinda Anantha Padmanabha}
\address{{\it Sibley School of Mechanical and Aerospace Engineering, Cornell University, Ithaca, NY 14850}}
\author{Jan Niklas Fuhg}
\address{{\it Department of Aerospace Engineering and Engineering Mechanics, The University of Texas at Austin, Austin, TX, 78712}}
\author{Cosmin Safta}
\address{{\it Sandia National Laboratories, Livermore, CA 94551}}
\author{Reese E. Jones}
\address{{\it Sandia National Laboratories, Livermore, CA 94551}}
\author{Nikolaos Bouklas$^*$}
\address{{\it  Sibley School of Mechanical and Aerospace Engineering, Cornell University \& Center for Applied Mathematics, Ithaca, NY 14850}}
\cortext[cor1]{Corresponding author: nb589@cornell.edu}
\begin{abstract}
Most scientific machine learning (SciML) applications of neural networks involve hundreds to thousands of parameters, and hence, uncertainty quantification for such models is plagued by the curse of dimensionality.
Using physical applications, we show that $L_0$ sparsification prior to Stein variational gradient descent ($L_0$+SVGD) is a more robust and efficient means of uncertainty quantification, in terms of computational cost and performance than the direct application of SGVD  or projected SGVD  methods.
Specifically, $L_0$+SVGD demonstrates superior resilience to noise, the ability to perform well in extrapolated regions, and a faster convergence rate to an optimal solution.
\end{abstract}

\begin{keyword}
Stein variational inference, projection, sparsification, uncertainty quantification, neural network, physical constraints, Bayesian neural network.
\end{keyword}
\end{frontmatter}

\section{Introduction}
Quantifying the uncertainty in the parameters of a model and thereby its predictions has become a central thrust in creating models for trustworthy simulation across engineering and science.
However, the well-established methods for obtaining posterior distributions of likely parameters such as Markov chain Monte Carlo (MCMC) sampling~\cite{neal1996monte} become infeasible with highly parameterized machine learning function representations, such as neural networks (NNs).
The \emph{curse of dimensionality} in this uncertainty quantification (UQ) setting is tied to the cost of sampling the posterior sufficiently to determine its covariance structure and generate representative push-forward realizations.

In this work, our goal is to obtain a high-dimensional posterior distribution over a large number of random variables representing model parameters, which is particularly useful when limited amount of training data is available.
One of the simplest ways to obtain the approximate posterior is to implement MCMC methods.
However, this approach is challenged by the number of parameters typically present in NNs and it is difficult to converge samples to those representative of the posterior, even for models with moderate dimensionality.
There has been enormous progress made to approximate high-dimensional posterior distributions using variational inference methods~\cite{zhang2018advances}.
However, these methods restrict the approximate posterior to a certain parametric family and find the best approximate posterior through optimization.
To surmount these issues, Liu and Wang~\cite{liu2016stein} recently proposed a non-parametric variational inference method called Stein variational gradient descent (SVGD).
Stein variational inference methods \cite{liu2016stein,leviyev2022stochastic}  and their projected variants~\cite{chen2019projected,chen2020projected} address shortcomings in both the reference standard MCMC methods, such as Hamiltonian Monte Carlo~(HMC)~\cite{betancourt2017conceptual}, and the ubiquitous \emph{mean field} variational inference technique~\cite{kurle2022detrimental,knoblauch2022optimization} by using an ensemble of particles that represent likely model realizations.
In SVGD, these particles simultaneously follow a gradient flow toward the true posterior which is augmented with repulsive forces that keep the realizations distinct.
Projected SVGD (pSVGD) starts with a model reduction step based on the Hessian at the maximum \aposteriori (MAP) estimate of the posterior distribution to divide the parameter space into an active component and an inactive complement.

Unlike pSVGD which relies on a subspace around the MAP, we propose that model parameter sparsification prior to uncertainty quantification can embed non-linear aspects of the reduction of a fully parameterized NN not captured in a linearization (Laplace-like approximation).
After regularization-based sparsification to obtain a reduced dimensionality parameter manifold, the proposed method proceeds with SVGD on the sparsified NN model.
We explore the $L_{p}$ family of regularizations including the recently introduced smoothed $L_0$ technique \cite{louizos2017learning}.
We focus this work on the uncertainty quantification of physical response models, specifically those that admit a potential and other structure, which additionally can be subject to a variety of constraints that we aim to strongly enforce.
To this end, we use combinations of sparsification and UQ methods, including pSVGD, full SVGD, $L_{0}$+HMC, and the proposed method ($L_0$+SVGD), to demonstrate their relative efficacy in this task.

In the next section, \sref{sec:related}, we give the background for the proposed methodology, followed by a description of the algorithms in \sref{sec:methods}.
In \sref{sec:results} we demonstrate these methods on two physical representation problems and compare results obtained via the competing algorithms.
Lastly, in \sref{sec:conclusion}, we conclude with a summary and directions for future work.

\section{Related work} \label{sec:related}

Our work draws on and we compare it to a number of uncertainty quantification, sparsification and machine learning techniques.

Variational inference (VI) is a UQ technique that recasts the UQ problem of constructing a posterior distribution of model parameters as an optimization problem.
VI fits a surrogate distribution, typically from a pre-selected family of distributions, to the available data through a Kullback-Liebler divergence measuring the similarity of the surrogate to the true posterior.
So-called \emph{mean field} VI limits the covariance of the surrogate to a diagonal matrix for computational efficiency and hence ignores parameter correlations.
Although widely used, this technique is known to generally underestimate uncertainty by construction due to the restricted covariance and the evidence lower bound objective \cite{blei2017variational}.
More recently, Liu and Wang \cite{liu2018stein} introduced Stein variational gradient descent which employs a coordinated ensemble of model realizations (\emph{particles}) to sample the covariance structure of the posterior.
Due to the limitations of applying this technique to models with a large number of parameters, subsequently Chen~\etal~\cite{chen2020projected} developed projected SVGD to handle parameter spaces that have an \emph{active} subspace of influential parameters.

Sparsification of model parameterizations has had a long history of development~\cite{bishop2006pattern,goodfellow2016deep}.
A primary method of sparsification is through regularization of the fitting objective by adding a secondary objective, which allows the model fit to compete with model complexity.
Willams~\cite{williams1995bayesian} introduced the $L_1$ regularization prior in the Bayesian setting that promotes sparsity due to the shape of the $L_1$ level sets.
Later, Louizos~\etal~\cite{louizos2017learning, louizos2017bayesian} introduced a practical $L_0$ regularization based on smoothing the counting norm.
$L_0$ regularization was employed in Fuhg~\etal~\cite{fuhg2024extreme} to great effect on physics-augmented models, which are the topic of this work.
In addition, Van Baalen~\etal~\cite{van2020bayesian} applied $L_0$ pruning in a Bayesian and precision quantization context for image classification.

Constraints on model structure, such as convexity, remove (parametric) model complexity that violates physical principles.
Amos~\etal~\cite{amos2017input} proposed the notion of a input convex neural network (ICNN), which embeds strict convexity in the model formulation.
This representation has been widely employed in the computational mechanics community in constructing well-behaved potentials~\cite{tac2022data,chen2022polyconvex,as2022mechanics,xu2021learning,klein2022polyconvex,klein2023parametrized,kalina2024neural,fuhg2022learning,fuhg2022machine} and other constructs such as yield functions~\cite{fuhg2022machine}.
Other properties such as positivity~\cite{xu2021learning,fuhg2023modular} and equivariance~\cite{ling2016machine,thomas2018tensor} can also be embedded in NN formulations.

\section{Methods}
\label{sec:methods}

When given data, it is standard practice in a Bayesian framework to assess the epistemic/reducible parametric uncertainty of a model by first finding a MAP estimate of the parameters and then using this estimate as the starting point for an MCMC sampling procedure~\cite{ghanem2017handbook} of the posterior distribution.
Unfortunately, for models with many parameters, such as NNs, the \emph{curse of dimensionality} prevents simple sampling methods from efficiently characterizing the distribution of likely parameters.
More efficient methods, such as those based on VI, have been developed to address this shortcoming.
In particular, Stein variational inference attains a degree of parallel efficiency by using a coordinated ensemble of model realizations (\emph{particles}) to explore the posterior distribution; however, the number of particles to fully characterize the posterior covariance still grows exponentially with the number of parameters.
As mentioned, pSVGD interleaves a model reduction step using a linear subspace arrived at through a proper orthogonal-like decomposition.
We propose to follow this notion by considering the MAP model structure and then applying Stein variational inference to this reduced dimensionality parameter manifold.
We believe this approach will be more effective in accommodating the complex nonlinear dependencies found in many NN models.
Of course, this depends on the effectiveness of the regularization scheme, as the premise is that a low-dimensional representation is sufficiently accurate for the physical problem.

\subsection{Bayesian calibration} \label{sec:bayes}

Given a dataset of input-output pairs $\data = \lbrace \inputvector^{i}, \outputvector^{i} \rbrace _{i=1}^{N}$, where $ N$ is the total number of training data and a model
\begin{equation} \label{eq:model}
\hat{\outputvector} = \NN(\inputvector; \parameters) \ ,
\end{equation}
Bayes rule provides a foundation for quantifying the uncertainty in the model parameters $\parameters$:
\begin{equation}\label{eq:post_main}
\prob(\parameters \giv \data) = \frac{\prob(\data \giv \parameters) \, \prob(\parameters)}{\prob(\data)}
\end{equation}
Here, the posterior $\prob(\parameters \giv \data)$ is proportional to the likelihood $\lik(\parameters) = \prob(\data | \parameters)$ multiplied by prior $\prob(\parameters)$, where the evidence $\prob(\data)$ is a constant, normalizing factor.
The MAP estimate $\parameters^*$ is a point estimate given by optimizing the log posterior
\begin{equation} \label{eq:log_post}
\parameters^* = \operatorname{argmax}_\parameters \left[ \log \prob(\parameters \giv \data) \right]
= \operatorname{argmax}_\parameters \left[ \log \prob(\data \giv \parameters)
+ \log \prob(\parameters) \right]
\end{equation}

In the absence of specific distributions for the discrepancy between the model and the data, we assume a multivariate normal distribution for the likelihood $\lik(\parameters)$, leading to
\begin{equation} \label{eq:log_post_mvn}
-\log \prob(\parameters \giv \data) =
\| \outputvector - \NN(\inputvector; \parameters) \|_{\Sigmab}^2 + \lambda \| \parameters \|_p
+ \text{constant} \ ,
\end{equation}
where $\Sigmab$ is the likelihood covariance that characterizes data noise and is usually taken to be diagonal.
In \eref{eq:log_post_mvn}, we also assumed specific forms for the prior distribution $\log \prob(\parameters)$ based on regularizing priors.
Hence the MAP can be obtained by the optimization of the loss $\loss$
\begin{equation}
\parameters^*
= \argmin_{\parameters} \underbrace{\left[
\| \outputvector - \NN(\inputvector; \parameters) \|_{\Sigmab}^2 + \lambda \| \parameters \|_p
\right]}_{\loss(\parameters; \data)}
\end{equation}
composed of the $\Sigmab$ weighted mean squared error with a secondary, complexity-reducing
regularization objective associated with the prior~\cite{williams1995bayesian,steck2002dirichlet,calvetti2018inverse}.

The connection between the MAP optimization loss and the log posterior shows how the prior can be identified with a penalization of non-zero parameters.
For instance, $L_2$ regularization is associated with a Gaussian prior
\begin{equation}
\prob(\parameters) = \Nc(\mathbf{0}, \sigma^2 \Ib; \parameters)
\propto \exp\left( - \frac{\| \parameters \|^2_2}{2\sigma^2} \right)
\end{equation}
where $\lambda = \sigma^{-2}$ acts as a penalty parameter in this context; likewise $L_1$ penalization corresponds to a Laplace prior
\begin{equation}
\prob(\parameters) \propto \exp\left( - \lambda \| \parameters \|_1 \right)
\end{equation}
The regularization norm in \eref{eq:log_post_mvn}, together with the likelihood, determines the sparsification pattern of the parameters in the NN model, \ie\ the \emph{active} and \emph{inactive} parameters.
The goal of sparsification is to reduce the number of parameters while maintaining accuracy.
This elimination of redundant parameters can promote generalization and will aid our goal of efficient and accurate uncertainty quantification.

A representation of the posterior itself can be obtained through sampling with MCMC methods like HMC \cite{neal1996monte} or with Stein variational inference methods like SVGD, which will be discussed in \sref{sec:SVGD}.
With a posterior on the parameters $\prob(\parameters \giv \data) $, we can then evaluate the pushforward distribution of the outputs by sampling the posterior and evaluating the model:
\begin{equation} \label{eq:pf}
\hat{\outputvector} = \NN(\inputvector; \parameters) \ \text{with}  \
\parameters \sim \prob(\parameters \giv \data) \ .
\end{equation}

\subsection{Smoothed $L_0$ sparsification} \label{sec:L0}

Sparsification by $L_0$ regularization employs the $L_0$ norm, also known as the \emph{counting} norm since it gives the cardinality of a set or vector, which is not differentiable.
The smoothed $L_0$ approach \cite{louizos2017learning} follows the general idea of a gating system where each trainable parameter is multiplied by a gate value $z\in[0,1]$, which makes the parameter inactive ($z=0$) or active ($z=1$).
The number of active gates, and therefore the model complexity, can then be penalized in the loss function.
However, due to the binary nature of the gates, this loss function is not differentiable.
Hence, following \cref{louizos2017learning}, we consider a reparametrization of the trainable parameters using a \emph{smoothed} gating system, i.e. let
\begin{equation}
\parameters = \overline{\parameters} \odot   \zs, \quad \text{with} \quad \zs = \min (\bm{1}, \max (\bm{0}, \overline{\bm{s}}))
\end{equation}
where $\odot$ denotes the Hadamard product and
\begin{eqnarray}
\overline{\bm{s}} &=& \bm{s} (\zeta- \gamma) + \gamma \bm{1},  \\
\bm{s} &=& \sig((\log \bm{u} - \log (1-\bm{u})+ \log \bm{\alpha})/\beta), \nonumber
\end{eqnarray}
Here, $\gamma$, $\beta$, $\zeta$ and $\log \alpha$ are user-chosen hyperparameters that define the smoothing of the vector of gate values $\zs$ and $\bm{u} \sim \Uc(\bm{0},\bm{1}) $ is a uniform random vector which is the same dimension as $\zs$.
Following the suggestions of \cref{louizos2017learning}, we set $\gamma = -0.1$, $\zeta = 1.1$, $\beta=2/3$, and obtain $\log \bm{\alpha} \sim \Nc(0,\sigma)$ by sampling from a normal distribution with zero mean and $\sigma=0.01$ standard deviation.
Since the gate vector is a random vector, we can define a Monte Carlo approximated loss function as
\begin{equation}
\begin{aligned}
\mathcal{R}(\overline{\parameters})&=\frac{1}{M} \sum_{j=1}^M\left(\frac{1}{N}\left(\sum_{i=1}^N \mathcal{L}\left(\NN
\left( \inputvector_i, \overline{\parameters} \odot \zs^m\right), \outputvector_i \right)\right)\right] & \\
& +\lambda \sum_{j=1}^\parameters \operatorname{sig}\left(\log \alpha_j-\beta \log \frac{-\gamma}{\zeta}\right) \ ,
\end{aligned}
\end{equation}
with $M$ being the number of samples of the Monte Carlo approximation and where $\lambda$ is the penalty weighting factor for the regularization analogous to that in \eref{eq:log_post_mvn}.
To make predictions at test time we can then set the values of the trainable parameters $\parameters^{*} = \overline{\parameters}^{*} \odot  \hat{\zs}$ where the gate values are obtained from
\begin{equation}
\hat{\zs} = \min ( \bm{1}, \max (\bm{0}, \sig(\log \bm{\alpha})(\zeta-\gamma)+ \gamma \bm{1})).
\end{equation}
\ie negative gate values correspond to inactive parameters.

\subsection{Stein variational inference} \label{sec:SVGD}

As mentioned in \sref{sec:bayes}, the posterior $\prob(\parameters|\mathcal{D})$ given in \eref{eq:post_main} can be sampled using MCMC methods like HMC.
However, with a large number of uncertain parameters $\parameters$, using traditional sampling methods can be a formidable challenge to obtain converged statistics due to sampling inefficiency, hyperparameter tuning, and the sequential nature of MCMC sampling.
On the other hand, most variational inference methods~\cite{blei2017variational} solve the Bayesian inference problem by minimizing the Kullback-Liebler (KL) divergence between the surrogate distribution $q(\parameters)$ and the posterior distribution $\prob(\parameters \giv \data)$.
The optimization problem is formulated in terms of the KL divergence as follows:
\begin{equation} \label{eq:surrogate_dist}
q^{*}(\parameters)
=\underset{q \in \mathcal{Q}}{\arg \min }\, {\KLdiv}(q(\parameters) \| \prob(\parameters | \mathcal{D}))
=\underset{q \in \mathcal{Q}}{\arg \min}\, \mathbb{E}_{q}[\log q(\parameters)-\log \tilde{p}(\parameters | \mathcal{D})] \ ,
\end{equation}
where $\tilde{p}(\parameters | \mathcal{D})= \prob(\mathcal{D} | \parameters) \prob(\parameters)=\prod_{i=1}^{N} \prob\left(\mathbf{y}^{i} | \parameters, \mathbf{x}^{i}\right) \prob(\parameters)$ is the unnormalized posterior.
The major drawback of this method is that it confines the approximate posterior to specific parametric variational families.

Therefore, we consider a non-parametric variational inference method called Stein variational gradient descent (SVGD) \cite{liu2017stein}.
This method initializes a set of \(S\) particles \(\{\boldsymbol{\parameters}_0^i\}_{i=1}^{S}\) that represent likely model parameterizations, and then iteratively moves them to the high posterior probability region using gradient information.
This update is guided by a step size and a perturbation direction given by the Stein discrepancy, ensuring the transformed particles align more closely with the target posterior, and can be interpreted \cite{liu2017stein} as a gradient descent algorithm, like Adam \cite{kingma2014adam}.
As in \cref{liu2016stein}, we employ the closed-form Stein discrepancy:
\begin{equation} \label{eq:stein_gradient}
\bm{\phi}^{*}(\parameters)
\propto \mathbb{E}_{\parameters^{'}\sim \mu}[\mathcal{T}^{\parameters^{'}}_{\prob}\kernel(\parameters,\parameters^{'})]
= \mathbb{E}_{\parameters^{'}\sim \mu}[\nabla_{\parameters^{'}}\log \prob(\parameters^{'} \giv \mathcal{D})\kernel(\parameters,\parameters^{'})+\nabla_{\parameters^{'}}\kernel(\parameters,\parameters^{'})],
\end{equation}
where \(\mathcal{T}_{\prob}\) is the Stein operator, \(\kernel(\boldsymbol{\parameters}, \boldsymbol{\parameters'})\) is a positive kernel, $\nabla_{\parameters^{\prime}} \log \prob\left(\parameters^{\prime}\giv \mathcal{D}\right) \kernel\left(\parameters, \parameters^{\prime}\right)$ is a kernel smoothed gradient, $\nabla_{\parameters} \kernel\left(\parameters, \parameters^{\prime}\right)$ is a repulsive force, and $\mu$ is the measure associated with the surrogate posterior distribution.
In this work, we choose a standard radial basis function kernel in the update procedure summarized in \Aref{alg:svgd}.
For more details on SVGD, see~\aref{app:SVGD}.

\begin{algorithm}[ht]
\textbf{Input}: \text{A set of initial particles} $\{\parameters_0^i\}_{i=1}^{S}$, \text{score function} $\nabla \log \prob(\parameters\mid \data)$, \text{kernel} $\kernel(\parameters,\parameters^{'})$, \text{step-size} $\{\epsilon_t\}$
\SetAlgoLined \\
\For{iteration t}{
$\bm{\phi}(\parameters_t^{i}) = \frac{1}{S} \sum_{j=1}^{S}\Big[\kernel(\parameters_t^j,\parameters_t^i)\nabla_{\parameters_t^j}\log \prob(\parameters_t^j, \mathcal{D})+\nabla_{\parameters_t^j}\kernel(\parameters_t^j,\parameters_t^i)\Big]$\\
$\parameters^i_{t+1}$ $\leftarrow$ $\parameters^i_t+\epsilon_t\bm{ \phi}(\parameters^i_t)$
}
\textbf{Result:} A set of particles $\parameters^i$ that approximates the target posterior
\caption{Stein variational gradient descent (SVGD)~\cite{liu2016stein}.}
\label{alg:svgd}
\end{algorithm}

The projected Stein algorithm~\cite{chen2020projected} propagates gradient descent on a subspace constructed from the likelihood Hessian at the MAP.
The subspace is constructed from the most significant generalized eigenvectors of the particle averaged Hessian.
The column matrix of the leading eigenvectors allows the projection of parameters into the active low-dimensional subspace where the likelihood informs the posterior
more than the prior does.
The full-dimensional posterior is then reconstructed by lifting the low-dimensional samples at a given gradient descent step and recombining them with the inactive part of the prior.
\Aref{alg:psvgd} summarizes the additional steps of pSVGD.

\begin{algorithm}[h]
\textbf{Input}: \text{A set of initial particles} $\{\parameters_0^i\}_{i=1}^{S}$, \text{score function} $\nabla \log \prob(\parameters\mid \data)$, \text{kernel} $\kernel(\parameters,\parameters^{'})$, \text{step-size} $\{\epsilon_t\}$
\SetAlgoLined \\
Form Hessian  $\Hs = \grad_\parameters \left[ \grad_\parameters \lik_\outputvector(\parameters)  \right]$ at the MAP $\parameters^*$.\\
Solve the eigenvalue problem $\Hs \psib_i = \lambda_i \Sigmab_0^{-1}\psib_i$
where $\Sigmab_0$ is the prior covariance. \\
Determine the active subspace of the $r$ eigenvectors  $\Psib = [\psib_0,\psib_1,\ldots,\psib_r]$ with spectral content 0.99 of the total \\
Construct the projector $\Ps = \Psib \Psib^T$ \\
\For{iteration t}{
$\bm{\phi}(\parameters_t^{i}) = \frac{1}{S} \sum_{j=1}^{S}\Big[\kernel(\parameters_t^j,\parameters_t^i)\nabla_{\parameters_t^j}\log \prob(\parameters_t^j, \mathcal{D})+\nabla_{\parameters_t^j}\kernel(\parameters_t^j,\parameters_t^i)\Big]$\\
$    \grad_{\parameters^r_j}\log \prob(\parameters^r_j \giv \data) = \Ps^T \grad_{\parameters_j}\log \prob(\parameters_j \giv \data) $ \\
$\parameters^i_{t+1}$ $\leftarrow$ $\parameters^i_t+\epsilon_t\bm{ \phi}(\parameters^i_t)$
}
Reconstruct $ \parameters^i = \Psib \parameters^i +  \parameters^* + \parameters^i_{\perp} $ where $\parameters^i_{\perp}$ are sampled from the prior and projected by the complement of $\Ps$\\
\textbf{Result:} A set of particles $\parameters^i$ that approximates the target posterior
\caption{Projected Stein variational gradient descent (pSVGD)~\cite{chen2020projected}.}
\label{alg:psvgd}
\end{algorithm}

\section{Results} \label{sec:results}

To ameliorate the curse of dimensionality and obtain the posterior distribution of model parameters, we propose first using $L_0$ sparsification to find a sparse model structure that approximates the MAP and then using SVGD on this compact parameterization ($L_0$+Stein).
We compare this methodology to SVGD alone (Alg. \ref{alg:svgd}) and pSVGD (Alg. \ref{alg:psvgd}) for UQ of physical NN models using examples from hyperelasticity and mechanochemistry.
For the SVGD methods, we explored $L_0$, $L_1$, and $L_2$ regularizations.
For these demonstrations, only $L_0$ regularization leads to parametrically compact models.
The other regularizations can reduce the total number of parameters, especially when combined with constraints on the weights as in ICNNs, but not as effectively as $L_0$.
For the compact $L_0$ hyperelastic model, which had less than 10 parameters, we were able to compare $L_0$+Stein to $L_0$+HMC results.

To each of the datasets we added multiplicative (heteroskedastic) noise
\begin{equation}
\outputvector = \etab * \hat{\outputvector}(\inputvector)
\end{equation}
where $\hat{\outputvector}$ is the output of the data generating model and $\etab \sim \Nc(\mathbf{0}, \sigma^2 \Ib)$ is independent, identically distributed Gaussian noise mimicking measurement noise.
We employed this noise to test the proposed method's prediction for out-of-training range where the noise level is different than in the training range.
Our comparison with HMC is the one exception where we added additive (homoskedastic) noise
\begin{equation}
\outputvector = \hat{\outputvector}(\inputvector) + \etab
\end{equation}
to connect to the classical UQ case \cite{kennedy2001bayesian}.

For all cases, we compare the push-forward posterior of the output $\outputvector$, as in \eref{eq:pf}.
We use the Wasserstein-1 ($W_1$) distance to compare these push-forward posteriors estimated by the competing methods to data and to each other
\begin{equation}
\text{W}_1(\prob_a(X), \prob_b(X)) = \int \vert \CDF_a - \CDF_b\vert \, \mathrm{d}X
\end{equation}
where the cumulative distribution functions (CDFs), associated with the probability density functions $\prob_a$, are defined empirically from the samples.
As a distance, a smaller $W_1$ implies that the two distributions are more similar.

\subsection{Hyperelasticity}
Hyperelasticity~\cite{fuhg2024extreme} assumes the existence of a potential $\Psi$ such that the stress $\Sb$ can be derived as
\begin{equation}
\Sb = 2 \partialb_\Cb \Psi,
\end{equation}
where $\Cb = \Fb^T \Fb$ is the left Cauchy-Green deformation tensor and $\Fb$ is the deformation gradient.
Assuming material isotropy implies that the invariants
\begin{equation}
I_1 = \tr \Cb, \quad
I_2 = \tr \Cb^*, \quad
J   = \sqrt{\det\Cb}
\end{equation}
fully determine $\Psi$, where $\Cb^* = \det(\Cb) \Cb^{-T}$ denotes the cofactor (adjugate) of $\Cb$.
Furthermore polyconvexity \cite{ball1976convexity,silhavy2013mechanics} requires that
$\Psi$ is convex in the three invariants and monotonically increasing in $I_{1}$ and $I_{2}$.

To embed polyconvexity and appropriately reduce the complexity of the potential stress response we use an  input convex neural network (ICNN)  \cite{amos2017input}, which is a modification of the well-known feedforward multilayer perceptron (MLP)~\cite{rosenblatt1961principles}.
In addition, we shift the potential to constrain the stress to be zero at the reference $\Fb=\Ib$:
\begin{equation} \label{eq:shift_potential}
\hat{\Psi} = \hat{\Psi}^{NN}(I_1,I_2,J)-\hat{\Psi}^{NN}(3,3,1)-\Psi^{S}(J),
\end{equation}
where $\hat{\Psi}^{NN}(I_1,I_2,J)$ and $\hat{\Psi}^{NN}(3,3,1)$ is the output from the NN, and $\Psi^{S}(J) = n(J-1)$ where $n$ is a constant that enforces stress normalization as in \cref{fuhg2023extreme}.
For this example, we consider a network architecture with  $2$ hidden layers, and 30 neurons in each hidden layer and Softplus activation functions.
Full details of the construction of an ICNN for this problem are given in \aref{app:icnn}.

We use the commonly employed Gent~\cite{ogden2004fitting,horgan2015remarkable} hyperelastic model for data generation.
It has a strain energy density
\begin{equation}
\Psi(I_1,I_2,J) = -\frac{\vartheta_{1}}{2} J_{m} \log \left( 1 - \frac{I_{1}-3}{J_{m}} \right) - \vartheta_{2} \log \left( \frac{I_{2}}{J} \right) + \vartheta_{3} \left( \frac{1}{2} (J^{2}-1) - \log J \right) ,
\end{equation}
with a complex dependence on the deformation invariants.
As in \cref{fuhg2023stress}, we chose the parameters to be  $J_{m}= 77.931$, $\vartheta_{1} =2.4195$,  $\vartheta_{2} =-0.75$ and $\vartheta_{3} =1.20975 $.
We observe the stress $\Sb$ for a uniform sampling of deformation space $[\defgrad]_{ij} \in \delta_{ij} + \Uc[-\epsilon,\epsilon]$  with $\epsilon=0.2$.
(Note $\det(\defgrad)$ is not controlled in this sampling scheme but $\det(\Cb)$ remains positive.)
We validate on a high symmetry, interpretable 1-parameter ($\gamma$) path through the reference configuration, namely constrained uniaxial extension
\begin{equation} \label{eq:validation_trajectory}
\Fb = \Ib + \gamma \eb_1 \otimes \Eb_1
\ \ \gamma \in [-0.4,0.4]
\end{equation}
with 1000 test points.
We use a mean square error loss $\Lc$ on the errors in the stress $\Sb$
\begin{equation}
\Lc = \frac{1}{N} \sum_i (\Sb_i - \hat{\Sb}(\Eb_i; \parameters))^2 + \lambda \| \parameters \|_p \ ,
\end{equation}
which we associate with the log posterior.
For all the cases, we employed the Adam optimizer~\cite{kingma2014adam} with the learning rate of $0.08$, $0.01$, and $0.005$ for $L_0$, $L_1$, and $L_2$ models.

\begin{figure}[H]
\centering
\includegraphics[width=0.31\textwidth]{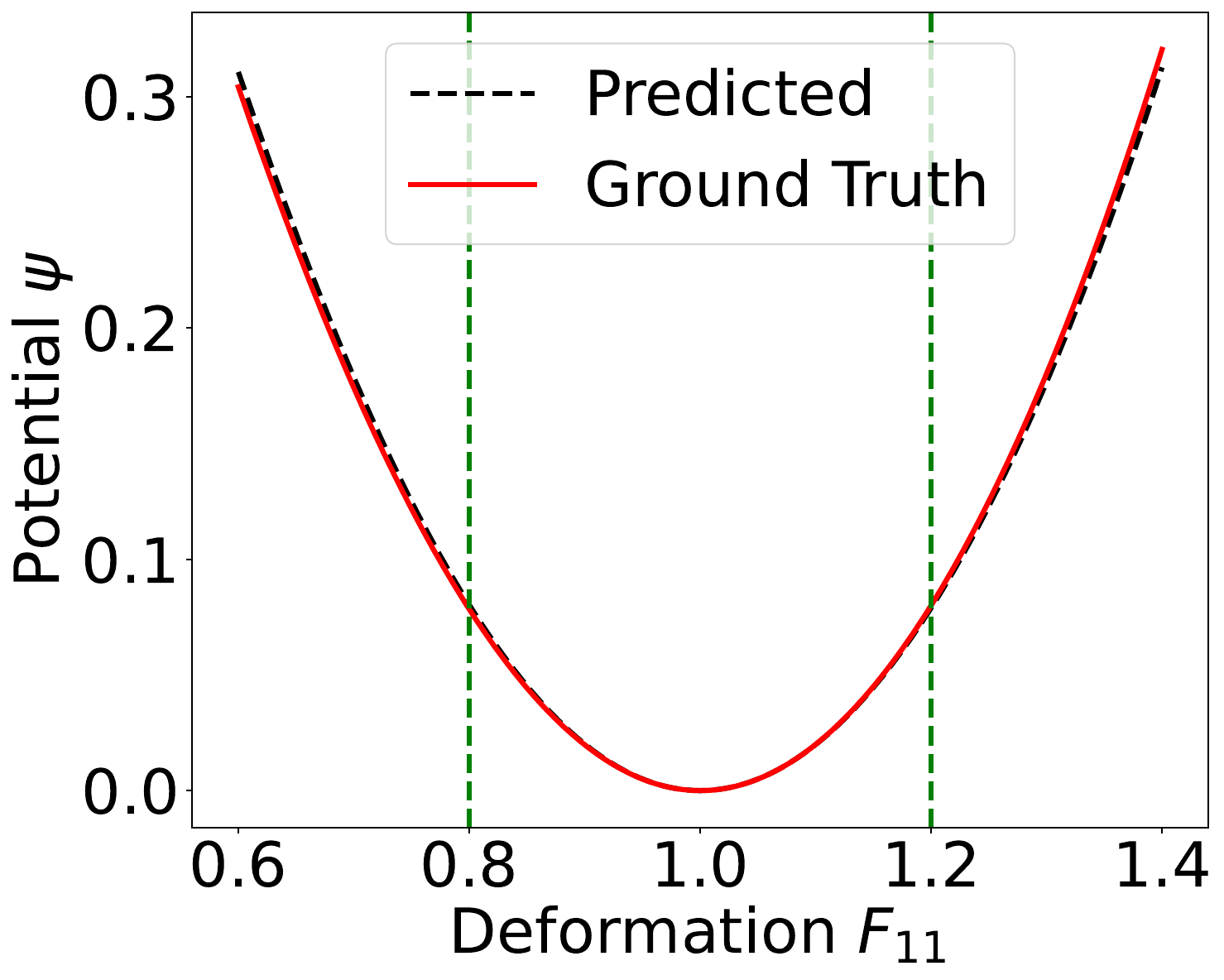}
\includegraphics[width=0.31\textwidth]{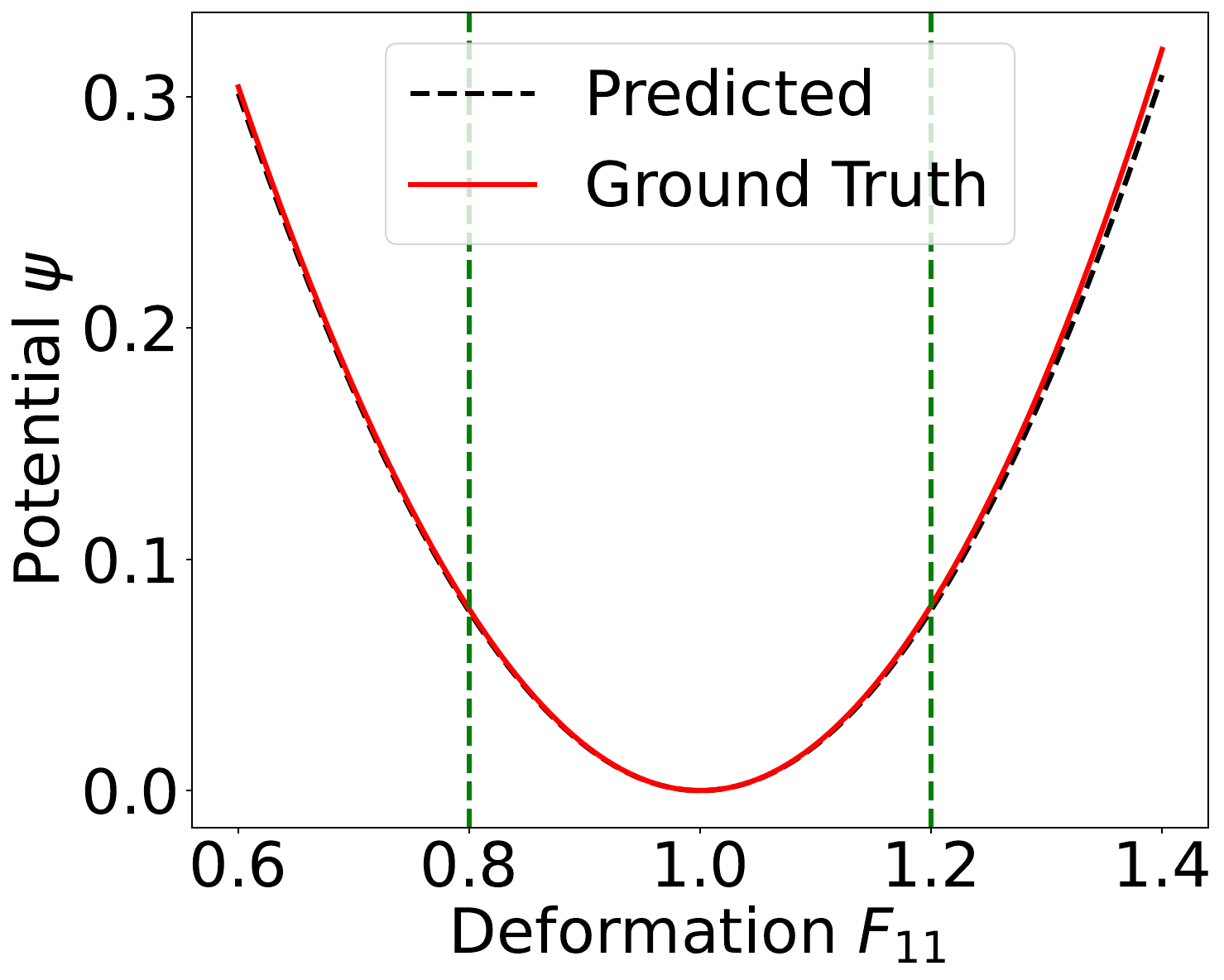}
\includegraphics[width=0.31\textwidth]{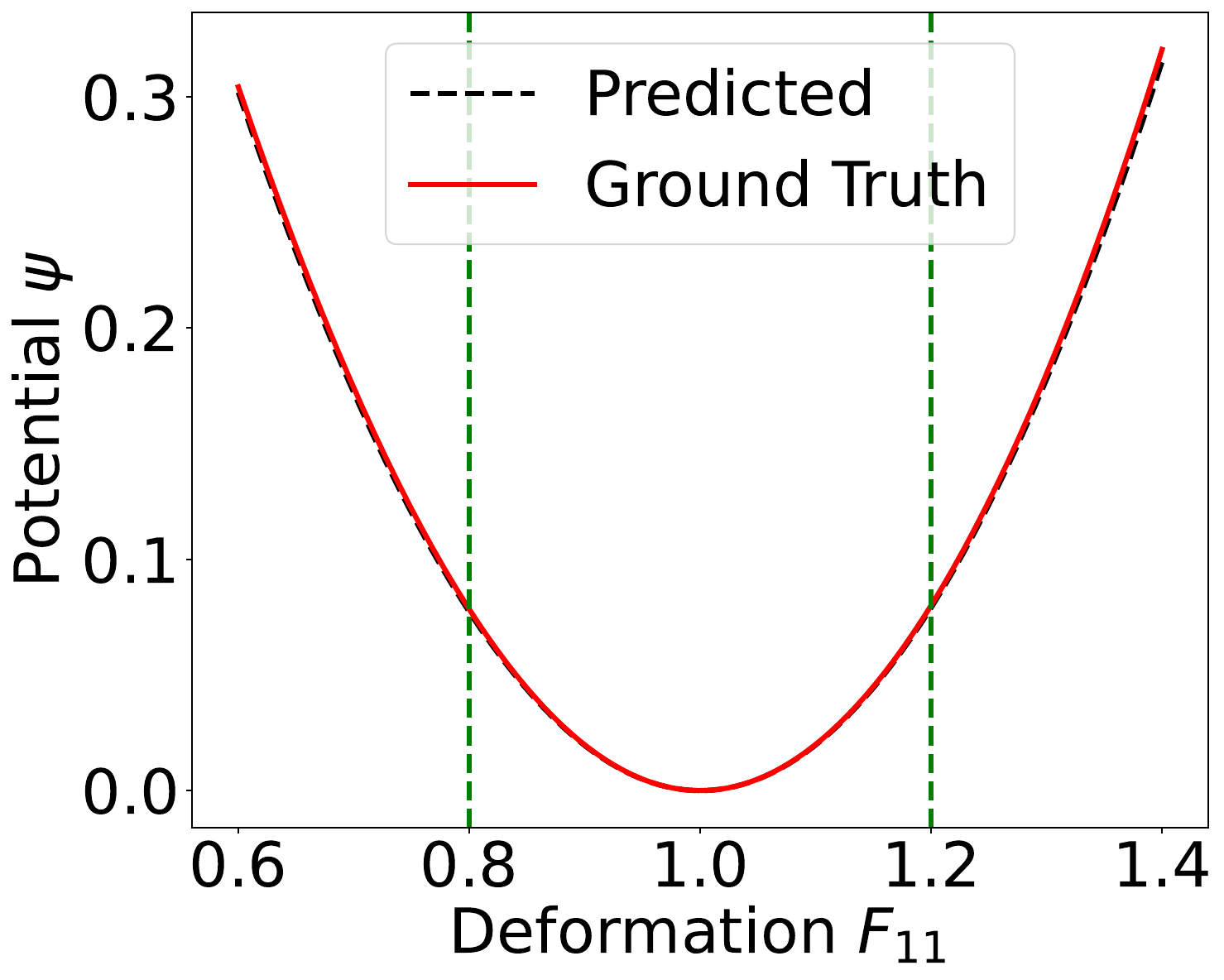}
\begin{minipage}{\textwidth}
\vspace{1cm}
\end{minipage}
\includegraphics[width=0.31\textwidth]{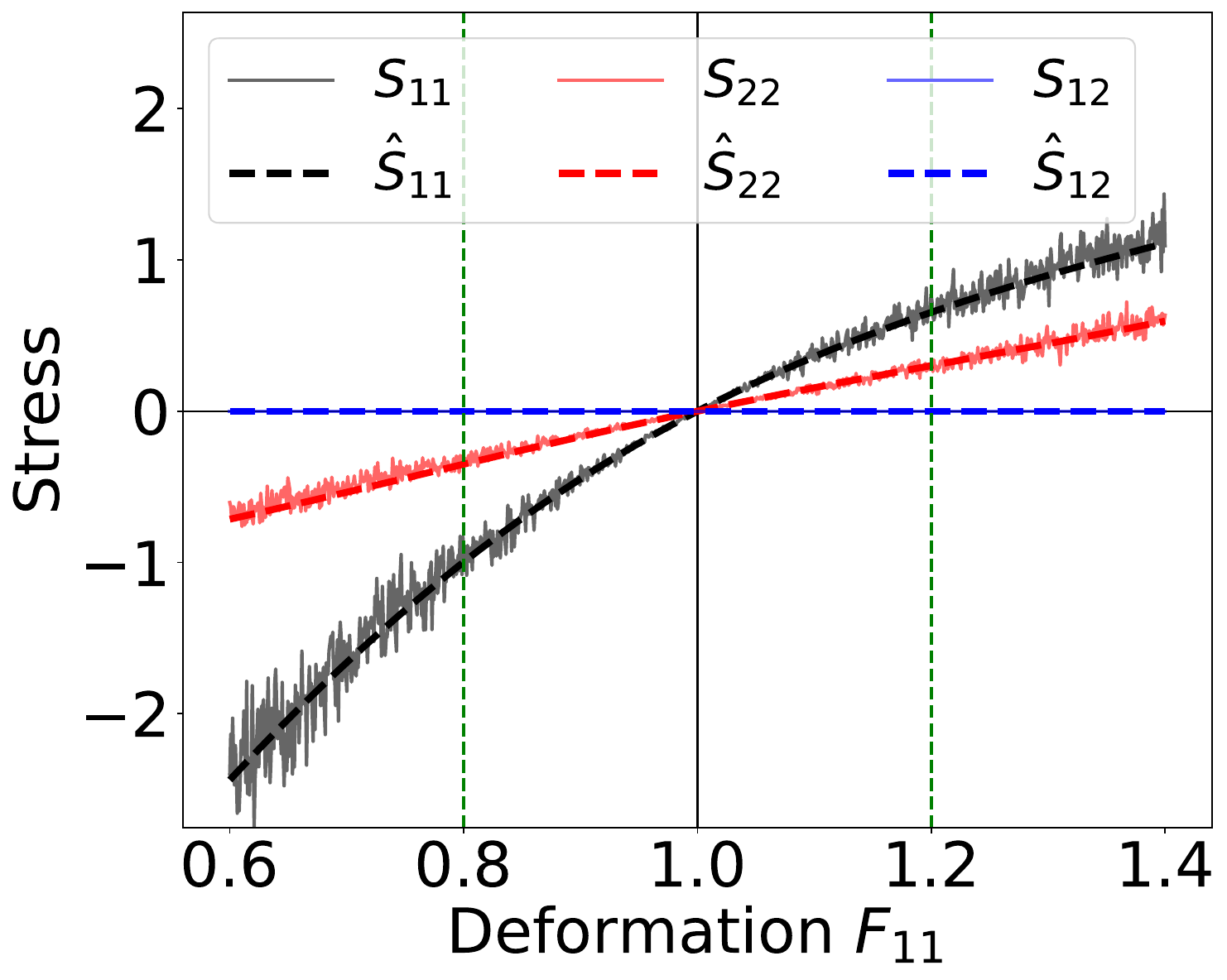}
\includegraphics[width=0.31\textwidth]{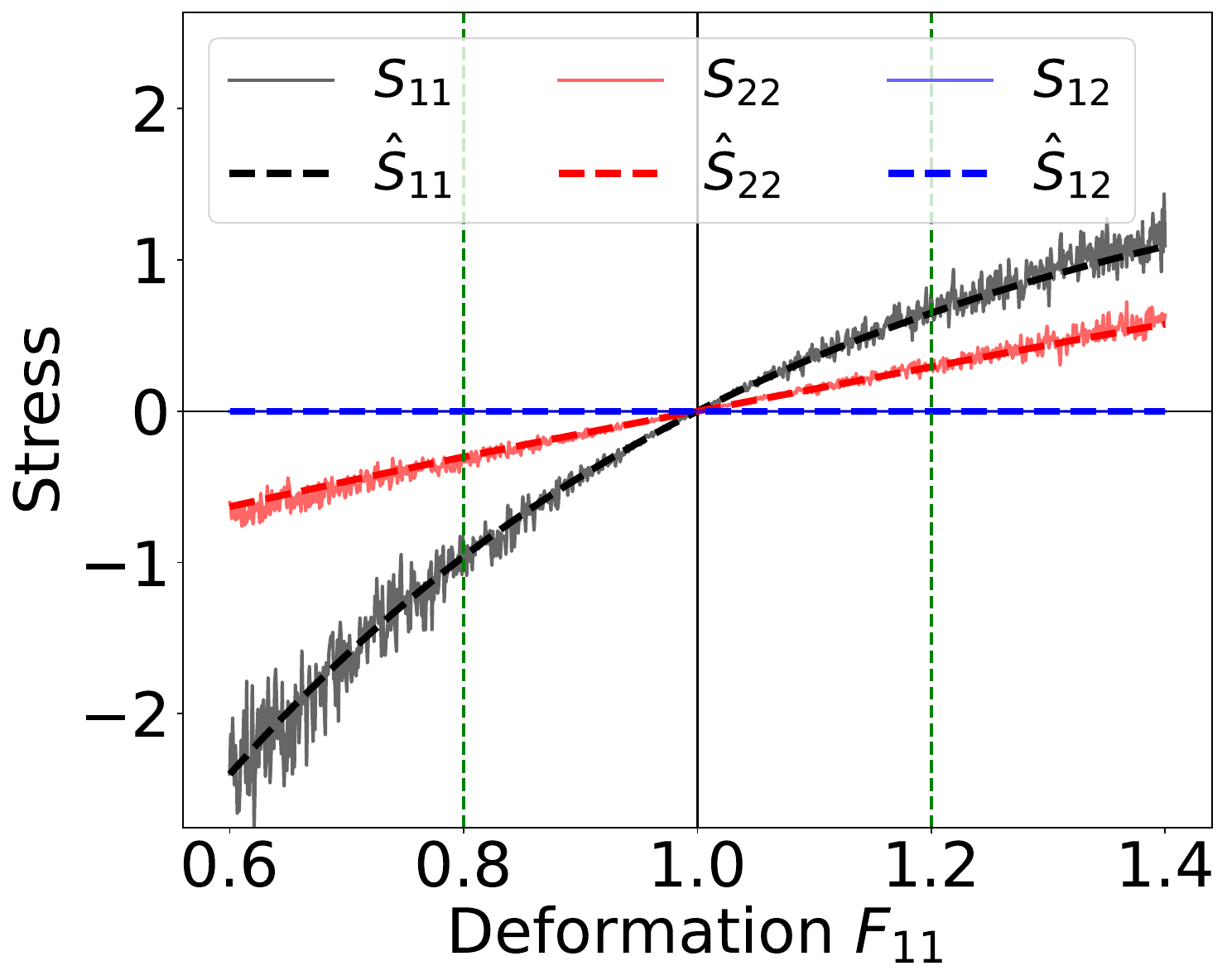}
\includegraphics[width=0.31\textwidth]{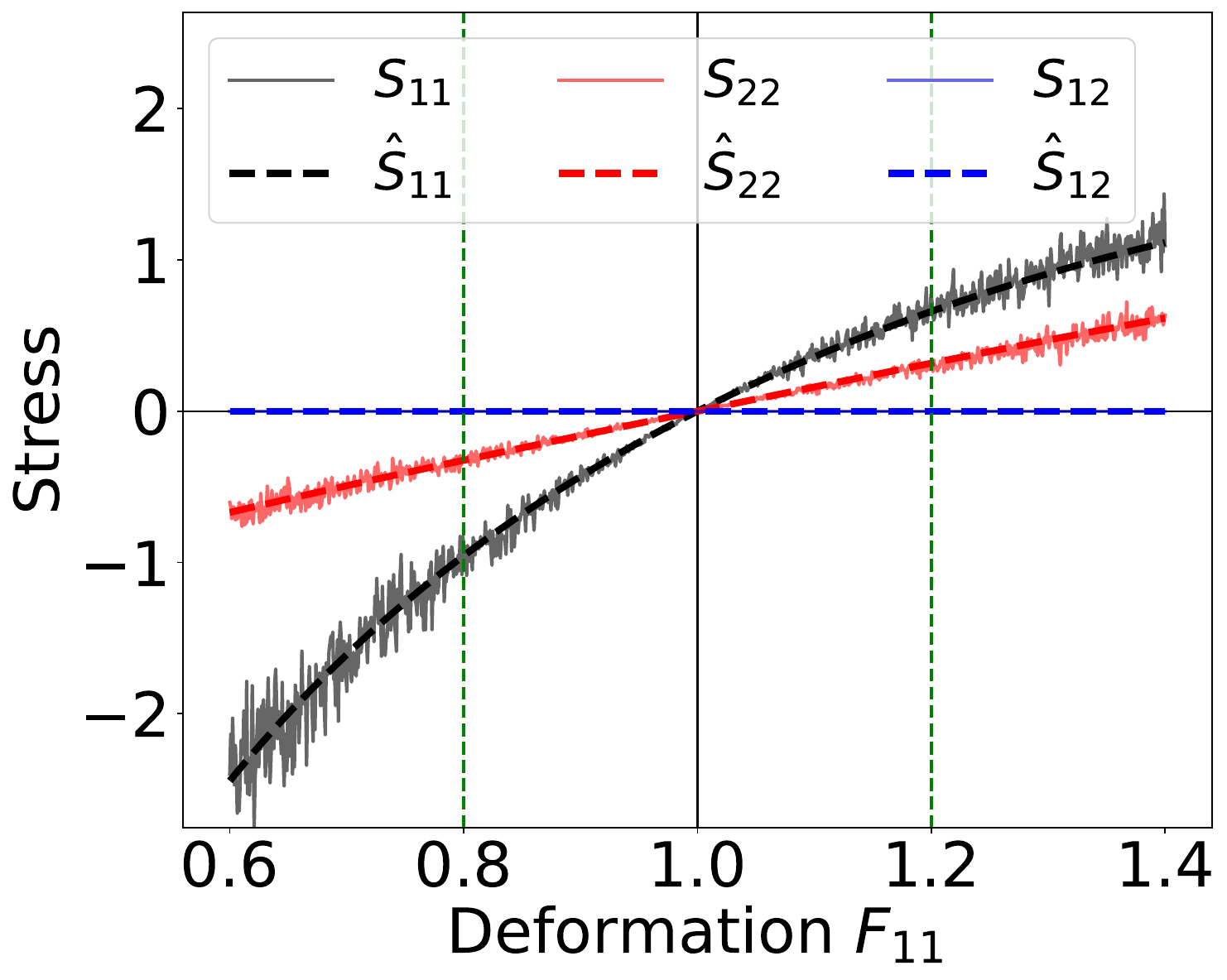}
\caption{Fits for $L_0$, $L_1$ and $L_2$ (left to right) regularization. Potential $\Psi$ (upper panels, compared to noiseless data) and stress $\Sb$ (lower panels, compared to 10\% noisy data). The total number of parameters for the  $L_0$, $L_1$ and $L_2$ fits are 7, 102 and 1005, respectively.
}
\label{fig:hyperelasticity_fits}
\end{figure}

\fref{fig:hyperelasticity_fits} shows the response of the MAP models on the validation data for $L_0$, $L_1$, and $L_2$ regularization.
The sequence of $L_2$, $L_1$, $L_0$ regularizations have an increasingly sharp tendency to promote sparsity.
Clearly, the $L_p$ MAP models are accurate in terms of the stress and the underlying potential in both interpolatory ( $0.6\leq F_{11} \leq 1.4$) and extrapolatory (outside the training data) regions, which are demarcated by the vertical green lines in \fref{fig:hyperelasticity_fits} and in subsequent figures.
In fact, the test $R^2$ score for all the three MAP models' is around $0.99$.
With $L_0$ sparsification, the MAP model with 7 parameters achieves an accuracy comparable to that of the $L_1$ and $L_2$ MAP models with 95 and 1005 parameters, respectively.
The $L_0$ representation is given by
\begin{align}
\hat{\Psi} &= 0.665 J + 5.623 \log{\left( \left(1 + e^{- 1.264 I_{2}}\right)^{0.764} \right. } \notag \\
&\quad \left. \left(e^{- 0.187 I_{2} - 0.339 J} + 1\right)^{1.8} e^{0.251 I_{1}} + 1 \right) - 9.71
\end{align}
The sparsification in $L_1$ and $L_2$ models is largely due to the weight clamping used to constrain weights to be positive in ICNN.

\begin{figure}[H]
\centering
\includegraphics[width=\figwidth]{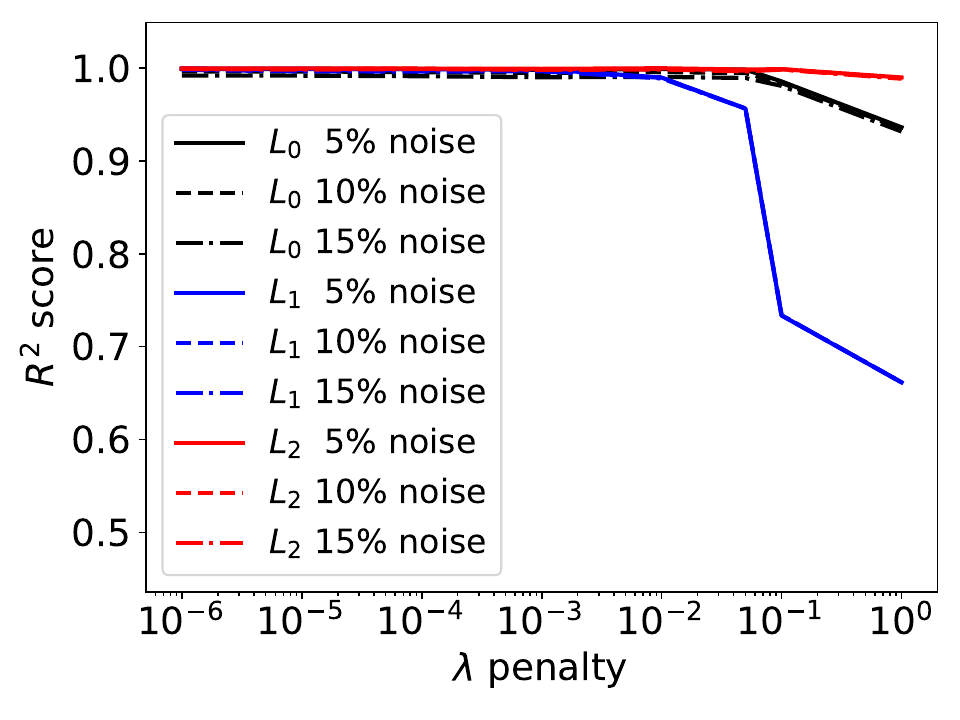}
\caption{Comparison of the L-curves for $L_{0}$, $L_{1}$ and $L_{2}$ regularizations and increasing amounts of additive noise using the test R$^{2}$ score.
The optimal penalty $\lambda$ depends strongly on the normalization but not as much on the added noise over the range that was studied.
}
\label{fig:hyperelasticity_Lcurves}
\end{figure}

We used classical L-curves~\cite{engl1994using} to determine the optimal penalty parameters $\lambda$.
\fref{fig:hyperelasticity_Lcurves} shows that rolloff in accuracy for each of the methods is distinct and hence indicates a well-defined optimal penalty, and that the optimal penalty is relatively insensitive to noise over the range we studied.

\begin{figure}[H]
\centering
\includegraphics[width=0.48\textwidth]{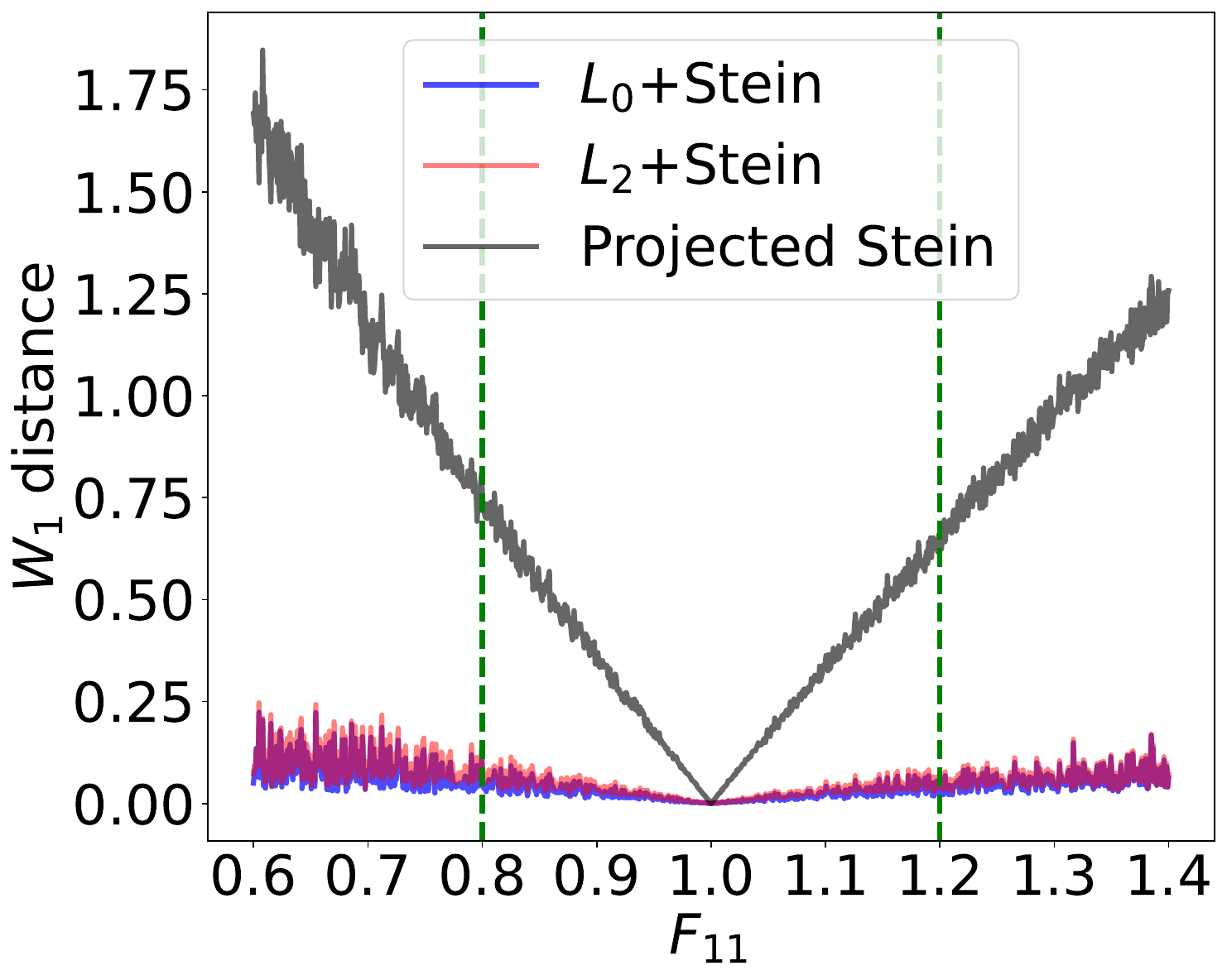}
\includegraphics[width=0.48\textwidth]{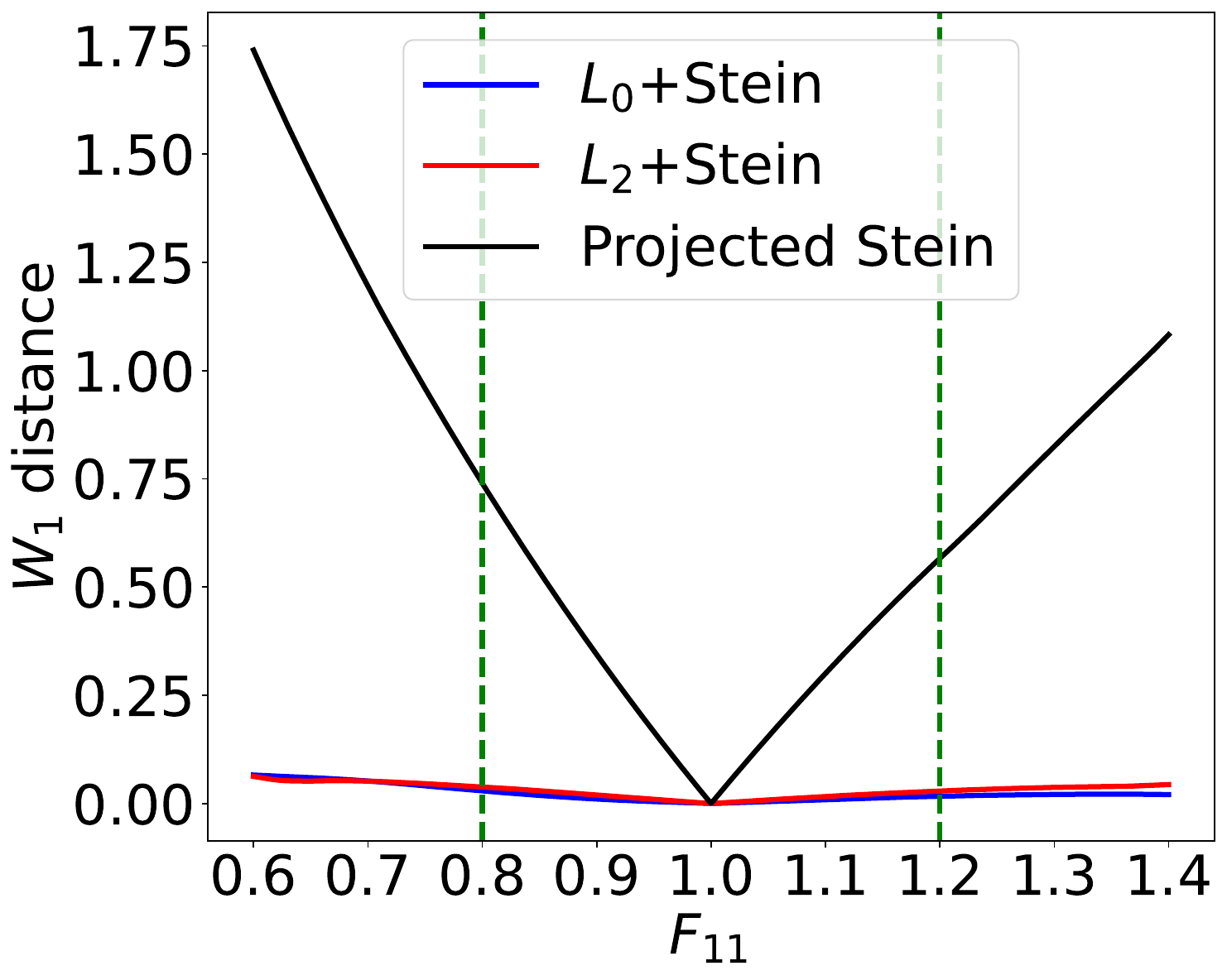}
\caption{Comparison of Wasserstein-1 distances between the $L_2$+Stein, $L_2$+projected Stein and $L_0$+Stein for noisy (left, 10\% heteroskedastic noise) and clean (right) data.
Note the similar trends for these two cases.
}
\label{fig:hyperelasticity_sparsification}
\end{figure}

\begin{figure}[H]
\centering
\includegraphics[width=\figwidth]{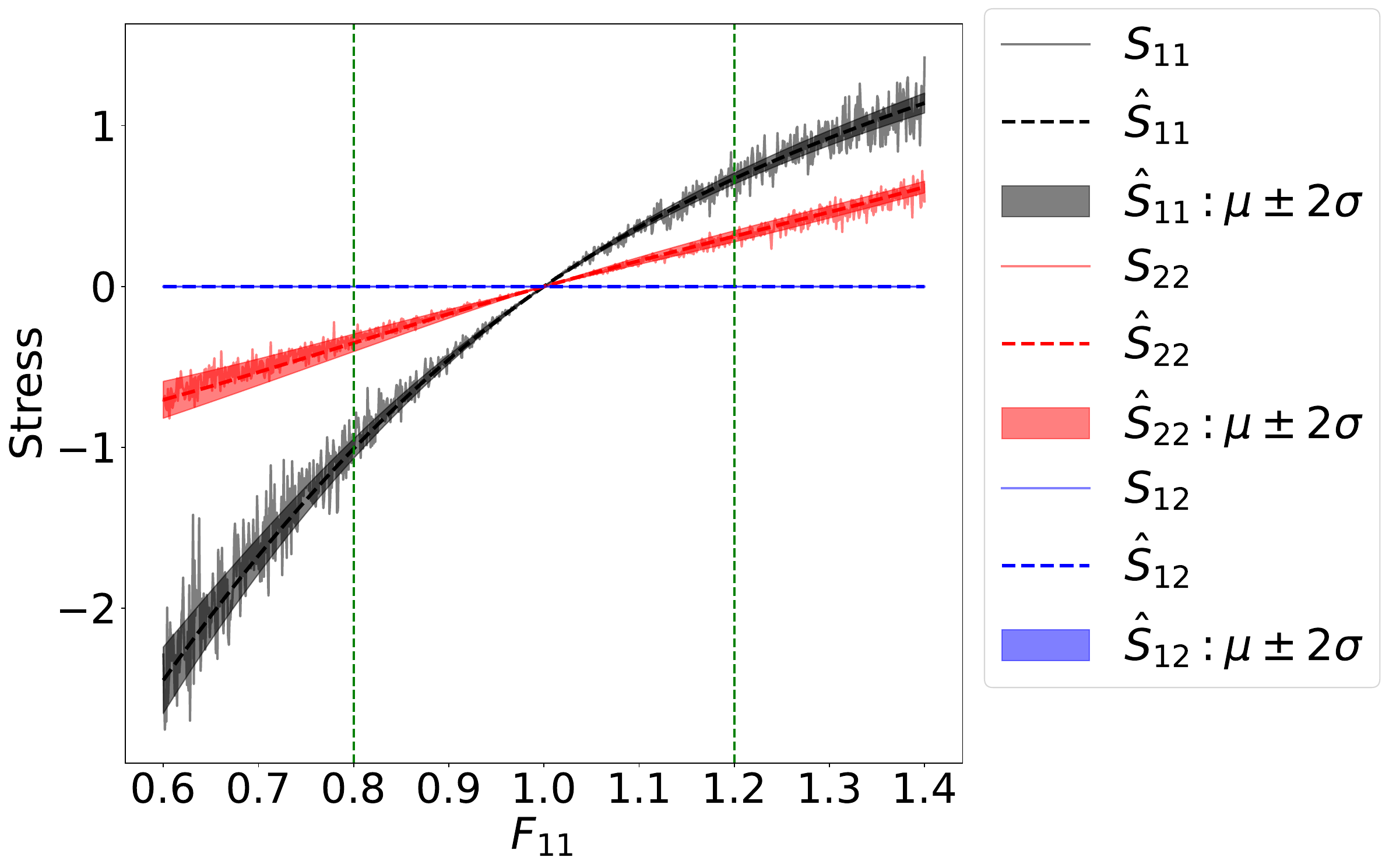}
\caption{Comparison of mean and standard deviation predicted by $L_0$+Stein pushforward samples and 10 \% heteroskedastic noisy validation data. Color bands indicate $\pm 2$ standard deviation from the mean, which largely overlap the noisy data.
}
\label{fig:hyperelasticity_uq}
\end{figure}

We compared the accuracy of $L_0$ sparsified SVGD to $L_2$ regularized SVGD and pSVGD in \fref{fig:hyperelasticity_sparsification} for both the noisy and clean data.
In both cases, clearly, the accuracy of pSVGD is relatively poor compared to the full SVGD methods for this example.
For this study we compare the predictions to noiseless data to show the small differences between the SVGD methods.
The sparsified $L_0$ model has marginally better accuracy than the $L_2$ model, which we attribute to the considerably smaller parameter space and, hence, the smaller dimensionality per particle.
\fref{fig:hyperelasticity_uq} illustrates how well the predictions of $L_0$+Stein match the held out noisy validation data.

\begin{figure}[H]
\centering
\includegraphics[width=\figwidth]{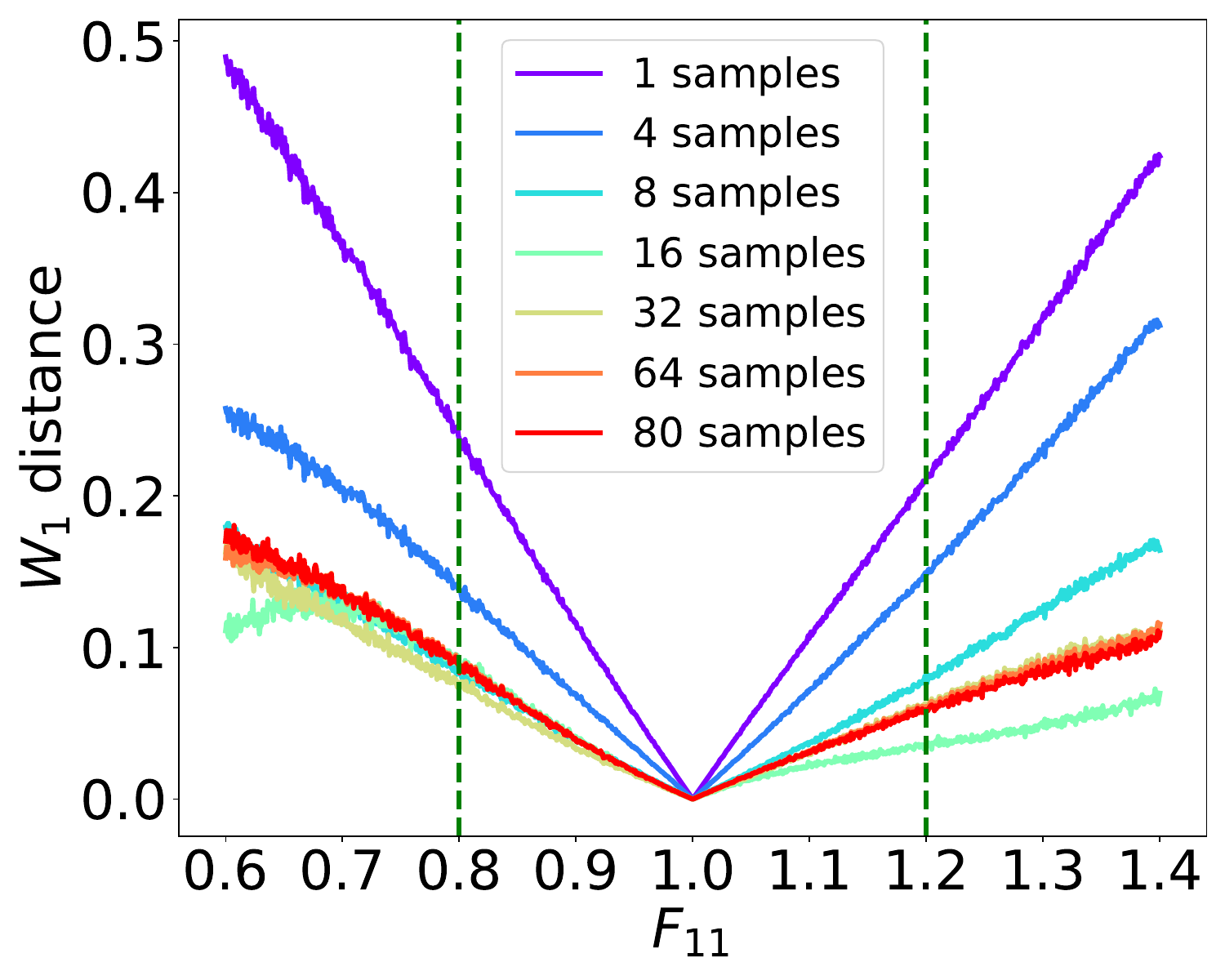}
\caption{Convergence of Wasserstein-1 distances for $L_0+$Stein results and the 10 \% heteroskedastic noise validation data distribution with an increasing number of data size $N_D$.}
\label{fig:hyperelasticity_epistemic}
\end{figure}

\begin{figure}[!htb]
\centering
\includegraphics[width=\figwidth]{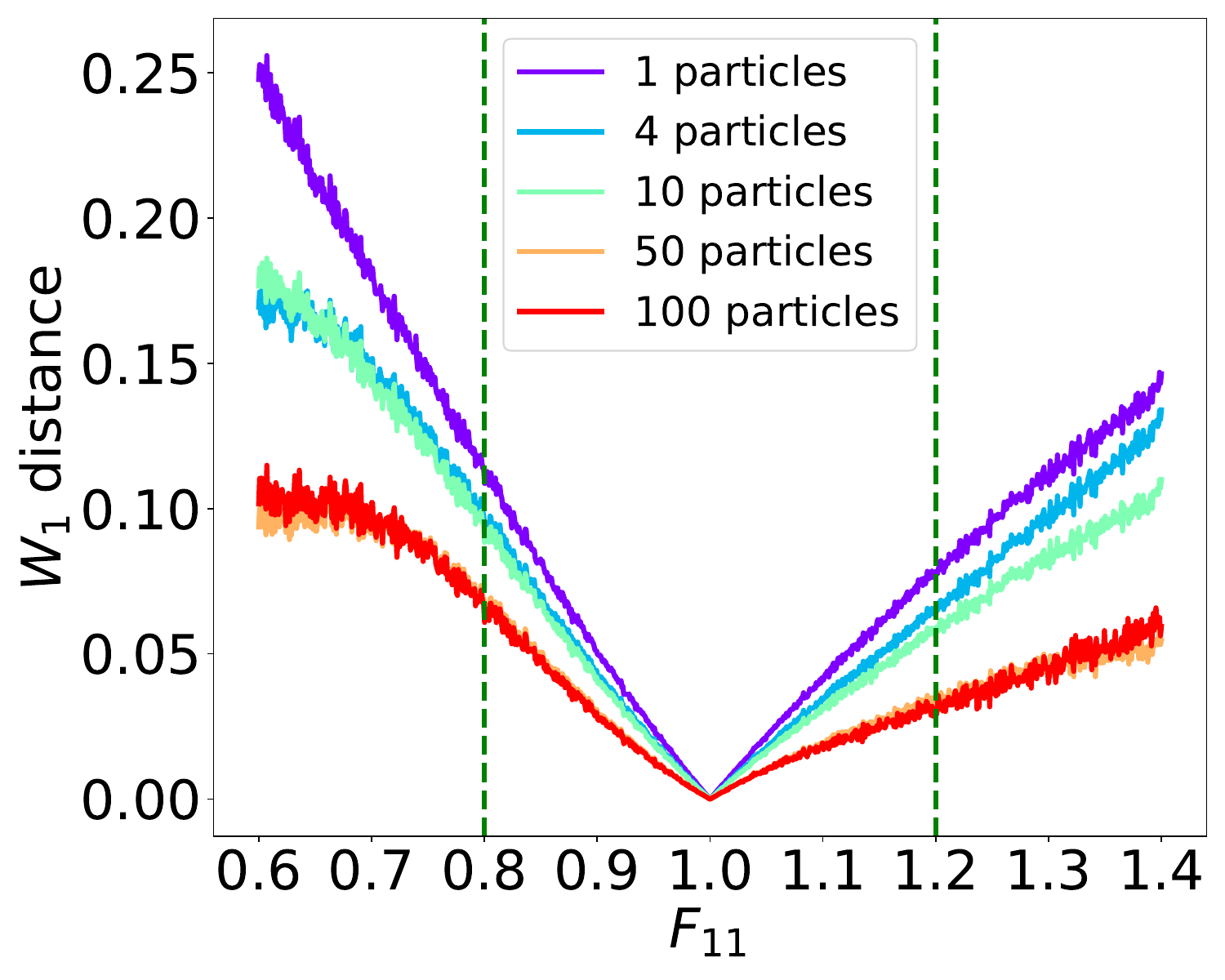}
\caption{Convergence of Wasserstein-1 distances for $L_0+$Stein results and the 10 \% heteroskedastic noise validation data distribution with an increasing number of particles. Note the 100 particle case overlaps the 50 particle result.}
\label{fig:hyperelasticity_particles}
\end{figure}

Now focusing on SVGD for the model structure obtained via $L_0$ sparsification, \fref{fig:hyperelasticity_epistemic} shows that the method converges to the validation data distribution with increasing amounts of data, albeit not entirely uniformly.
Furthermore, the general trend of the distribution similarity metric $W_1$ with deformation $F_{11}$ is plausible.
It is zero at the reference where data and the model are constrained to be zero with certainty.
The $W_1$ distance grows quasi-linearly and asymmetrically from this reference point as does the stress response.
For this study, we used $10$ particles.
Likewise, \fref{fig:hyperelasticity_particles} shows that the SVGD method also converges to the validation data distribution for increasing ensemble size.
For this study, we used 80 training data.
Also, from \tref{tab:mechchem_cost}, we observe that the computational cost for the $L_0$+Stein approach is significantly lower than that of $L_2$+Stein approach for this numerical example.

\begin{figure}[!htb]
\centering
\includegraphics[width=\figwidth]{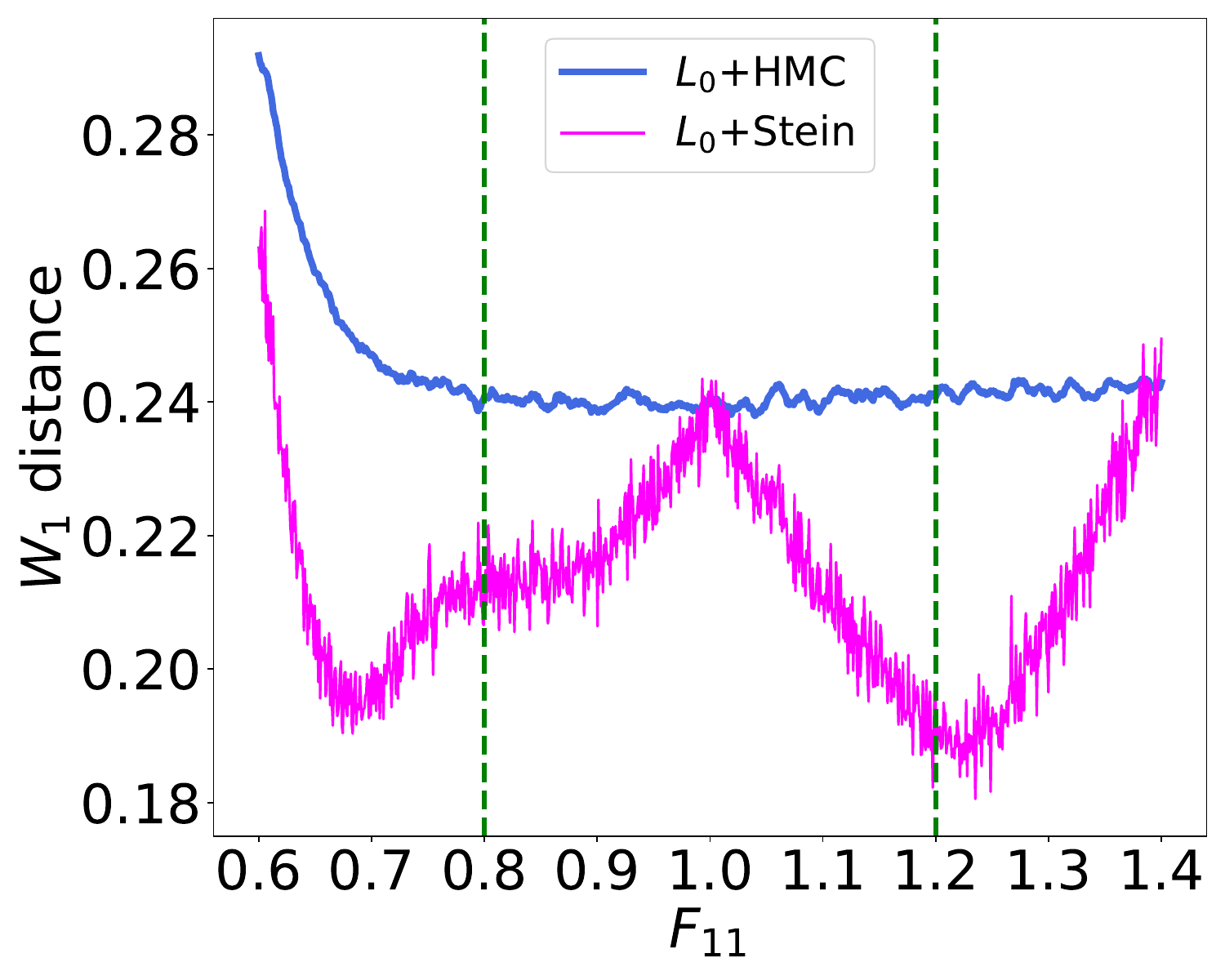}
\caption{Comparison of Wasserstein-1 distances between the push-forward distributions obtained by SVGD and HMC, respectively and the 10\% homoskedastic noise validation data.
}
\label{fig:hyperelasticity_hmc}
\end{figure}

Lastly we compare $L_0$+Stein with $L_0$+HMC applied to the same model using 80 samples and 10\% homoskedastic noise.
\fref{fig:hyperelasticity_hmc} shows that SGVD is closer to the validation data distribution despite using a HMC chain with $10^5$ steps (decimated to $10^3$ samples).


\subsection{Mechanochemistry}
\label{sec:mechchem}
\begin{figure}[H]
\centering
\includegraphics[width=1\textwidth]{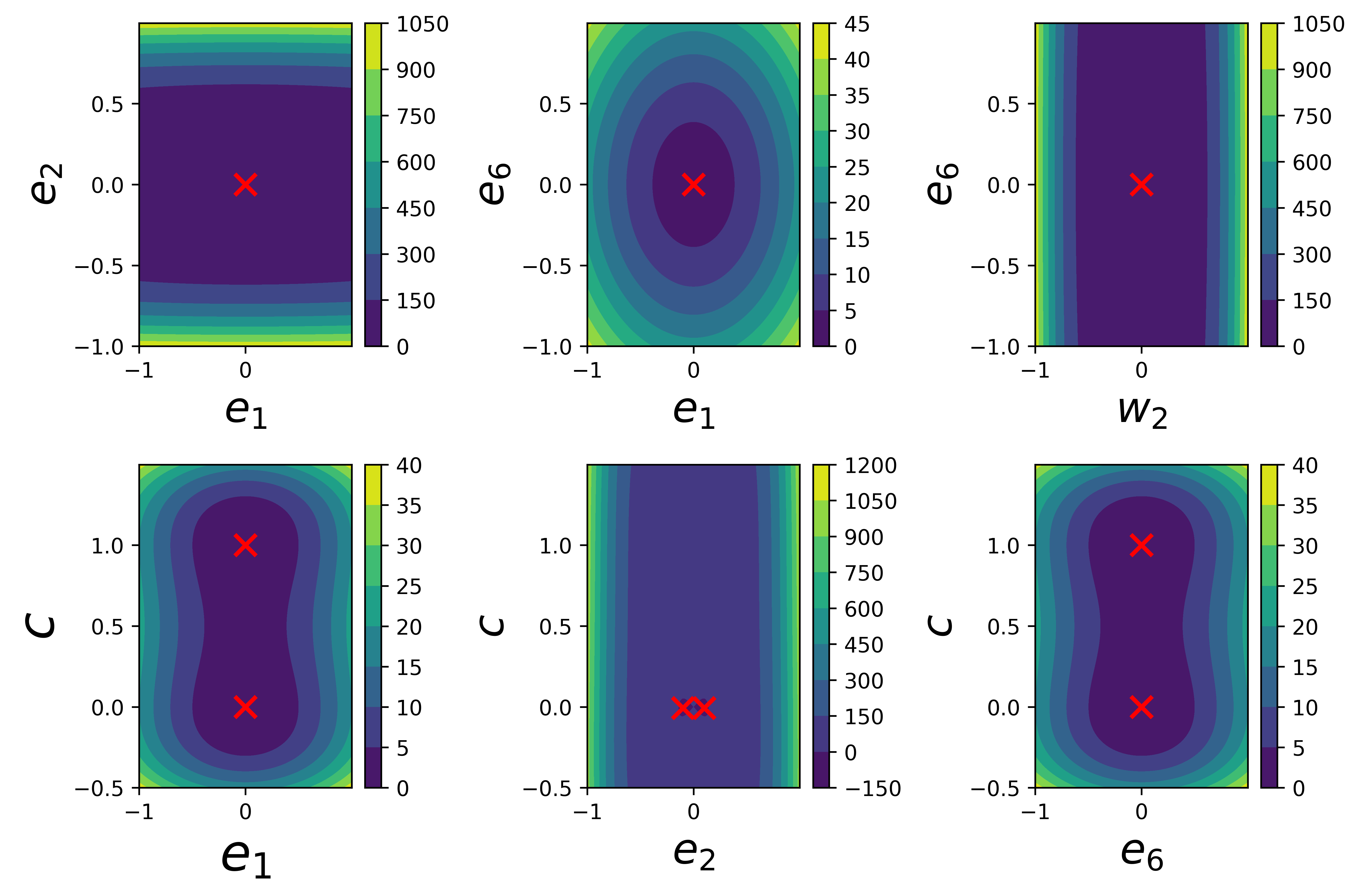}
\caption{Free energy potential, slices through the reference configuration $\Fb=\Ib$.
Red $\Xs$'s mark the location of the minima of the potential in these slices.
}
\label{fig:mechchem_potential}
\end{figure}

In phase change processes, a free energy potential $\Psi$ dependent on deformation and chemical concentration plays a central role~\cite{garikipati2017perspectives,rudraraju2016mechanochemical,teichert2023bridging}.
In this demonstration, we take the form of the free energy from~\cref{rudraraju2016mechanochemical}:
\begin{equation}
\begin{aligned}
\Psi(\Eb,c) = & 16 d_c c^4-32 d_c c^3+16 d_c c^2 +\frac{2 d_e}{s_e^2}\left(e_1^2+e_6^2\right) +\frac{d_e}{s_e^4} e_2^4+(2 c-1) \frac{2 d_e}{s_e^2} e_2^2
\end{aligned}
\end{equation}
where
$d_c = 2.0$, $d_e=0.1$, $s_e = 0.1$,
and
\begin{align}
e_1 &= \frac{1}{\sqrt{3}} \tr \Eb, &
e_2 &= \frac{1}{\sqrt{2}}(E_{11}-E_{22}), &
e_6 &= \sqrt{2} E_{12}, &
e_3 = e_4 = e_5 = 0
\end{align}
and $\Eb = 1/2 (\Cb - \Ib)$ is the Lagrange strain and $c$ is the concentration.
The free energy has multiple wells which are characteristic of a material that can undergo a phase change and this potential is highly nonlinear and non-convex as the projections in \fref{fig:mechchem_potential} show.
Note we use a 2D, plane strain reduction of $\Eb$.
Here, we enforce the normalization condition for the free energy by setting:
\begin{equation}
\hat{\Psi} = \hat{\Psi}^{NN}(e_1,e_2,e_6,c)-\hat{\Psi}^{NN}(0,0,0,0).
\end{equation}
in a manner similar to \eref{eq:shift_potential}.
For this demonstration, we observe  the stress
\begin{equation}
\Sb \equiv \partial_\Eb \Psi
\end{equation}
and the chemical potential
\begin{equation}
\mu \equiv \partial_c \Psi \ .
\end{equation}
for a uniform sampling of deformation space $[\defgrad]_{ij} \in \delta_{ij} + \Uc[-\epsilon,\epsilon]$  with $\epsilon=0.2$ and concentration $c \in \Uc[0,1]$.
The loss $\Lc$ balances the errors in the stress $\Sb$ and the chemical potential $\mu$
\begin{equation}
\Lc = \sum_i (\Sb_i - \hat{\Sb}(\Eb_i,c_i; \parameters))^2
+ \sum_i (\mu_i - \hat{\mu}(\Eb_i,c_i; \parameters))^2
+ \lambda \| \parameters \|_p
\end{equation}
We use same path for validation as in \eref{eq:validation_trajectory} augmented with the linear path $c=1.25(\gamma+0.4)$ with $\gamma \in [-0.4,0.4]$ through $c \in [0,1]$.

For this example, we consider a NN model with the input being the Lagrange strain and concentration, and the model predicts the free energy $\Psi$ with three hidden layers with $4,16,4$ hidden units, respectively, and \emph{softplus} activations.

\begin{figure}[H]
\centering
\includegraphics[width=0.32\textwidth]{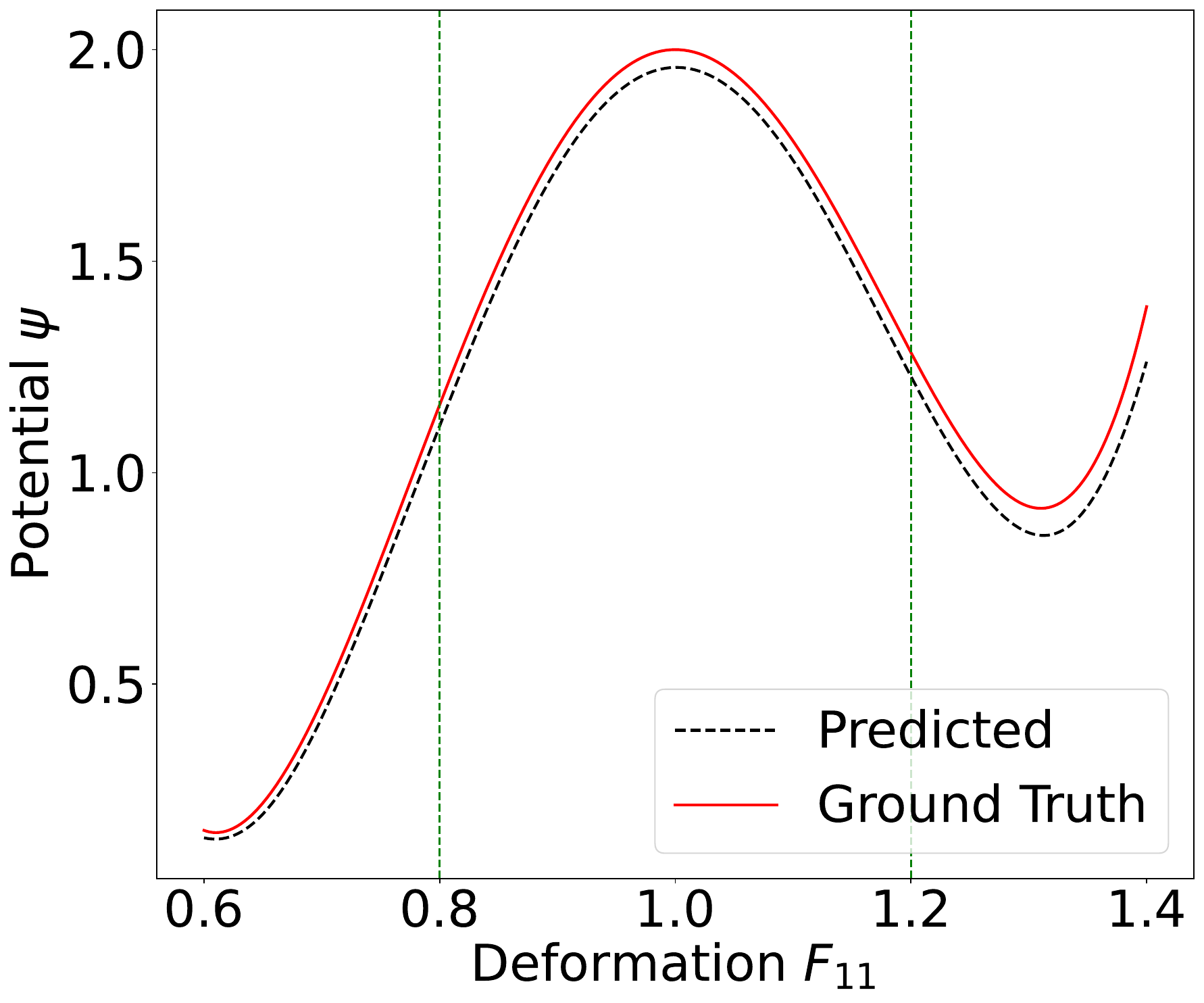}
\includegraphics[width=0.32\textwidth]{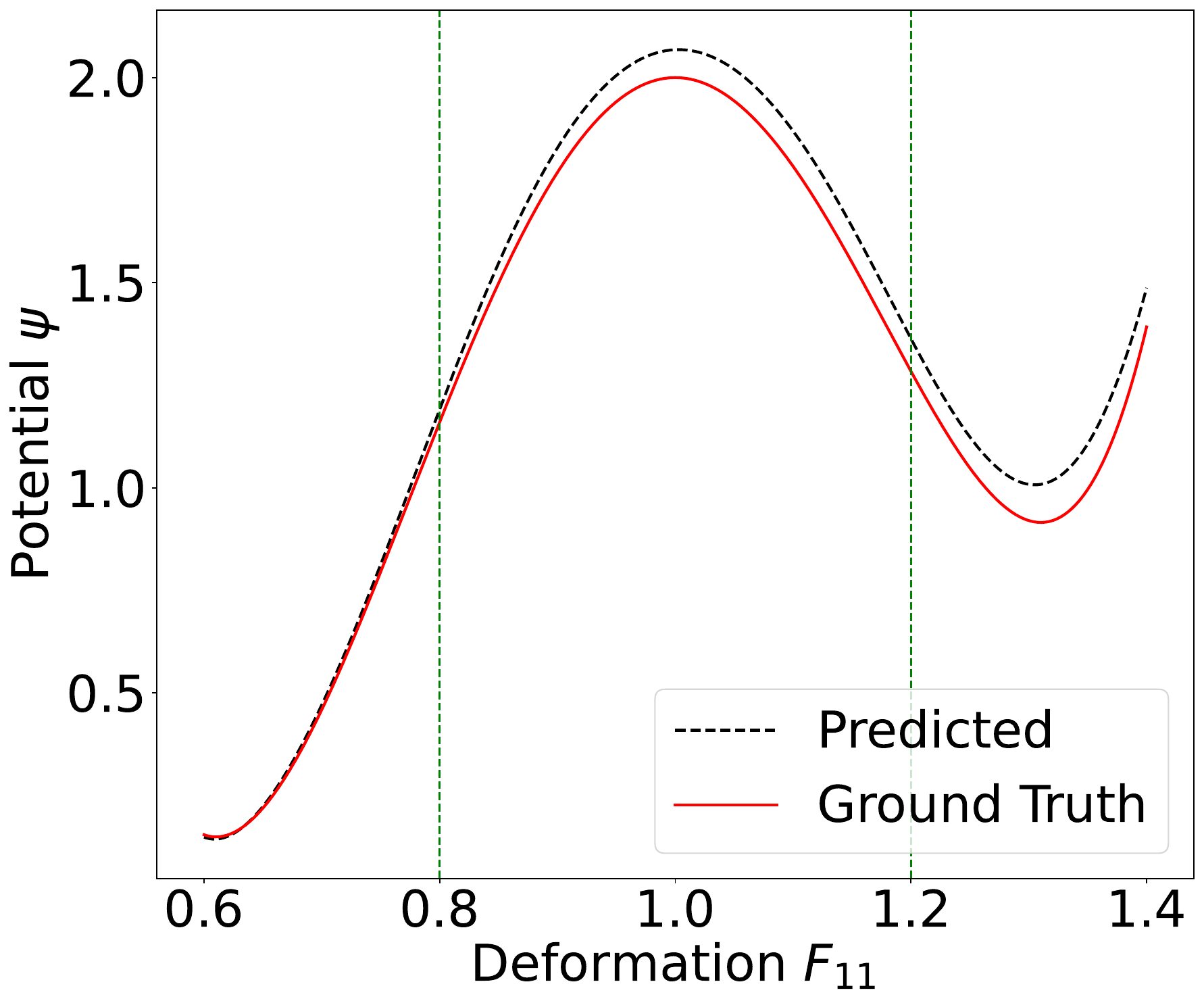}
\includegraphics[width=0.32\textwidth]{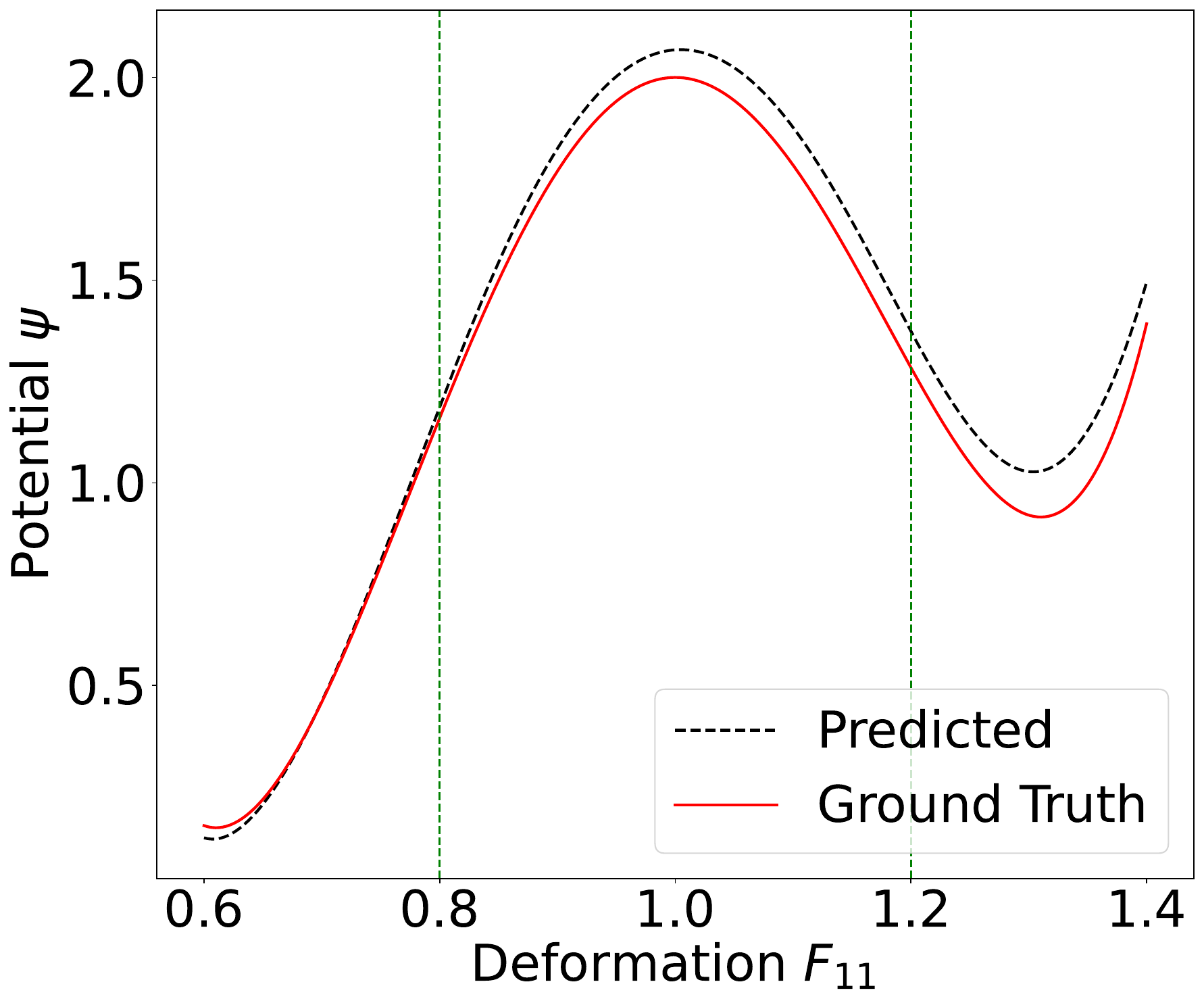}
\begin{minipage}{\textwidth}
\vspace{1cm}
\end{minipage}
\includegraphics[width=0.32\textwidth]{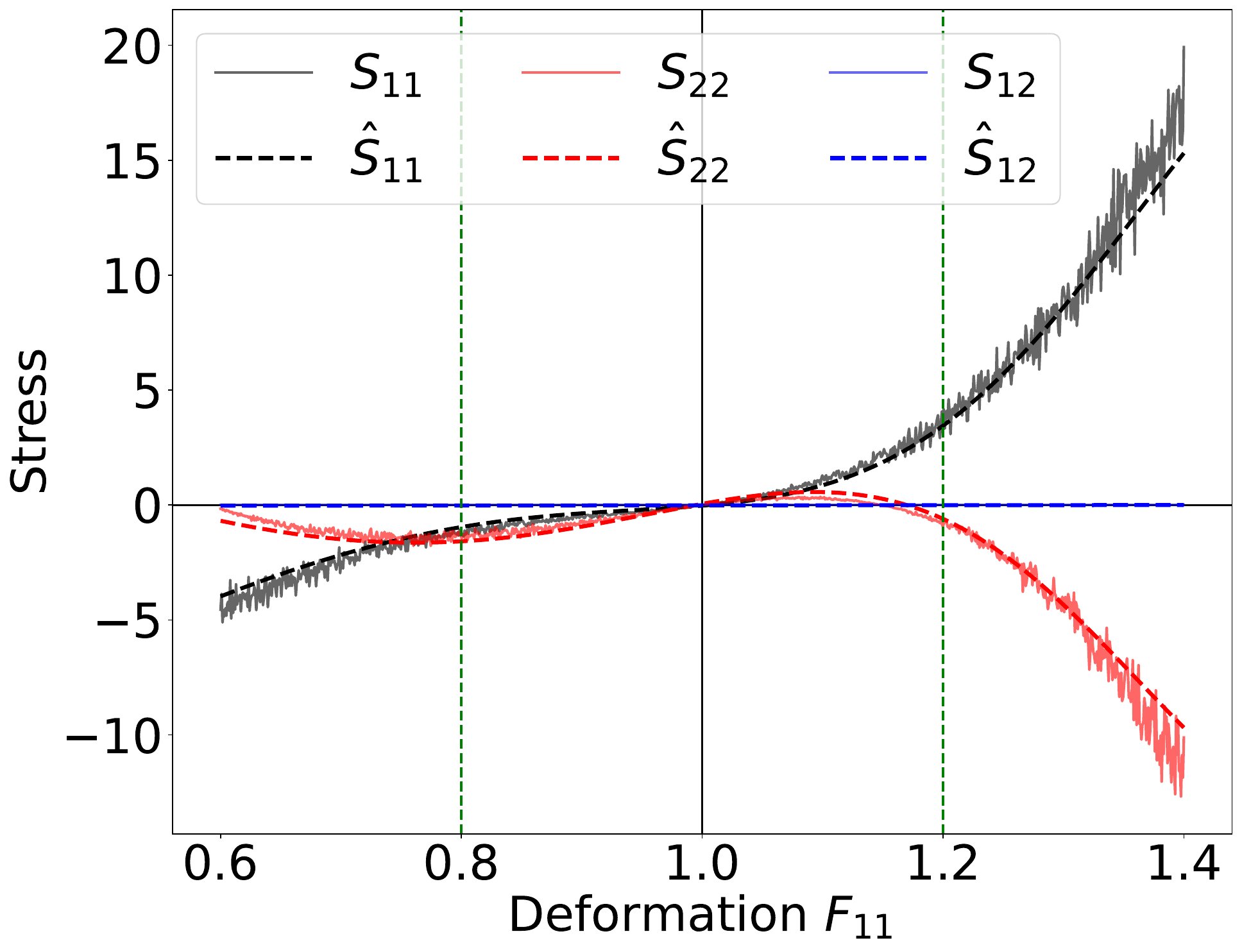}
\includegraphics[width=0.32\textwidth]{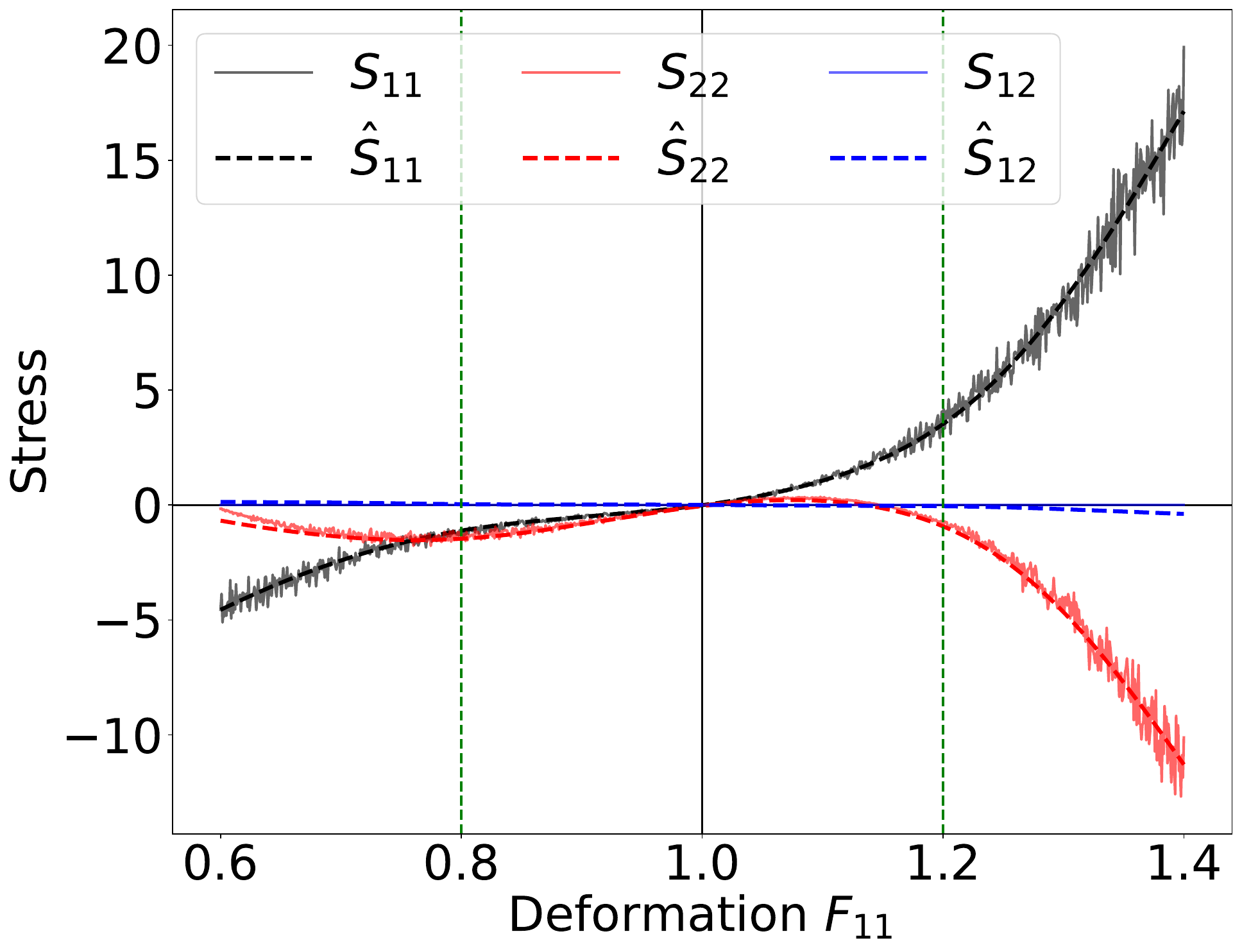}
\includegraphics[width=0.32\textwidth]{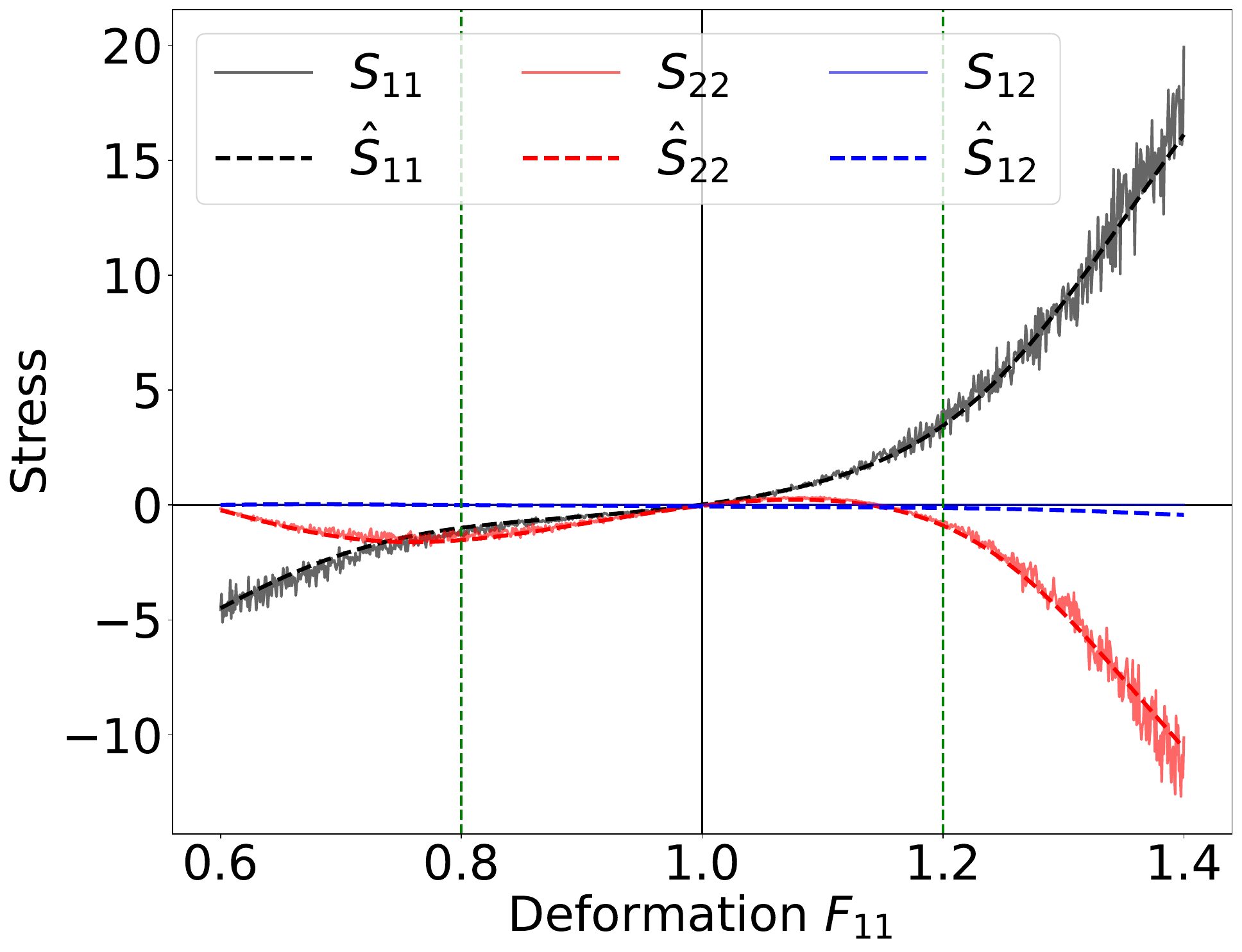}
\begin{minipage}{\textwidth}
\vspace{1cm}
\end{minipage}
\includegraphics[width=0.32\textwidth]{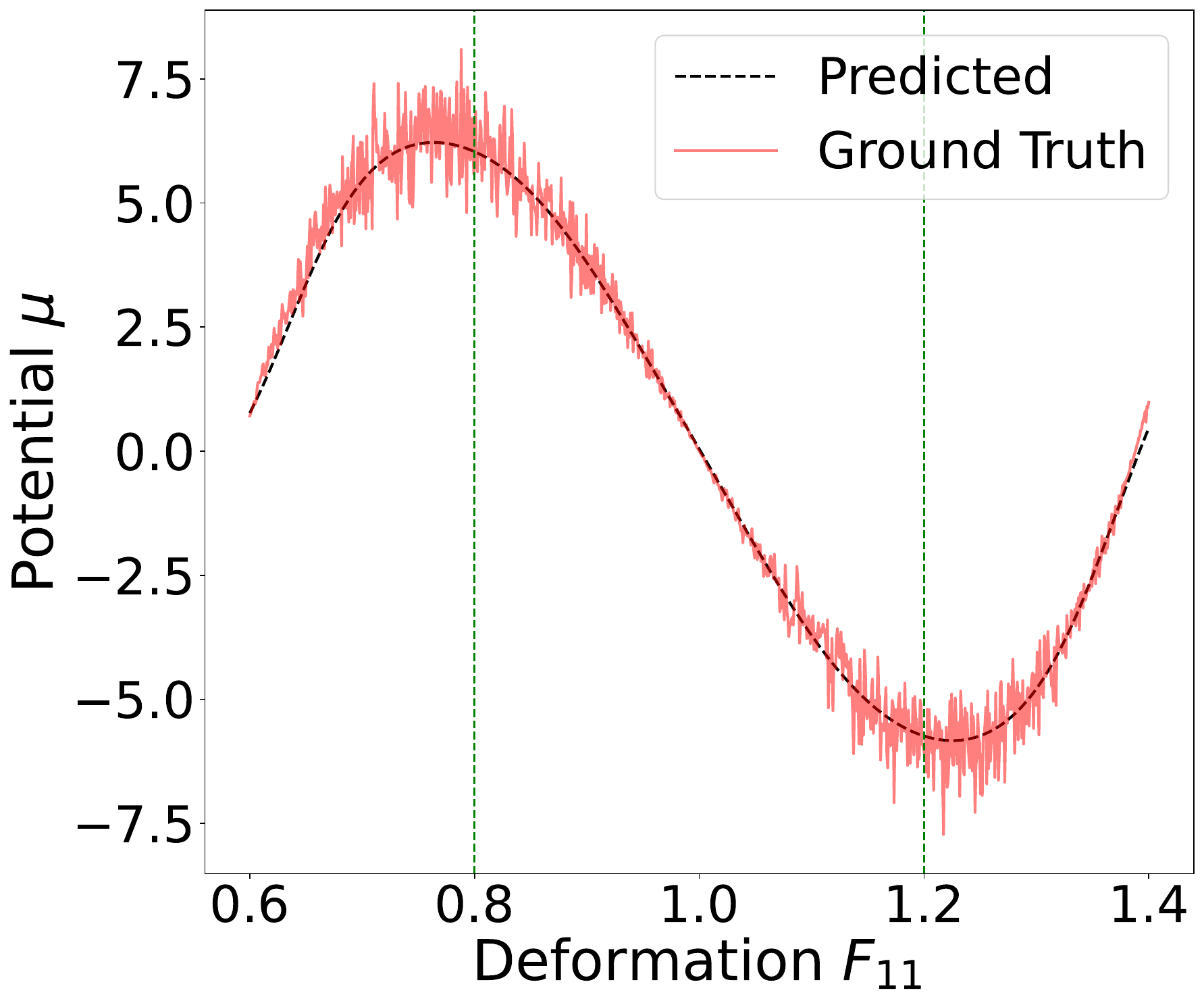}
\includegraphics[width=0.32\textwidth]{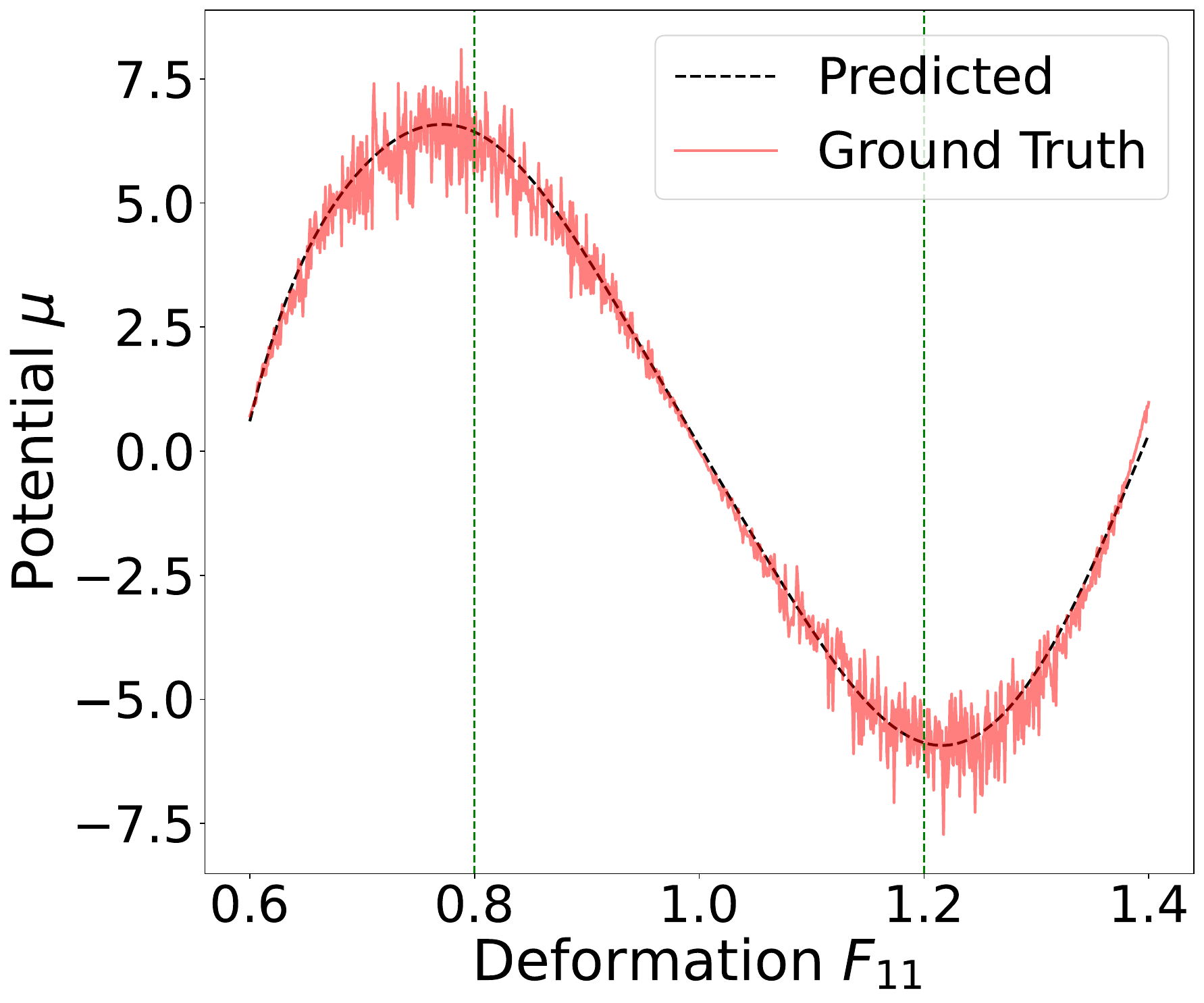}
\includegraphics[width=0.32\textwidth]{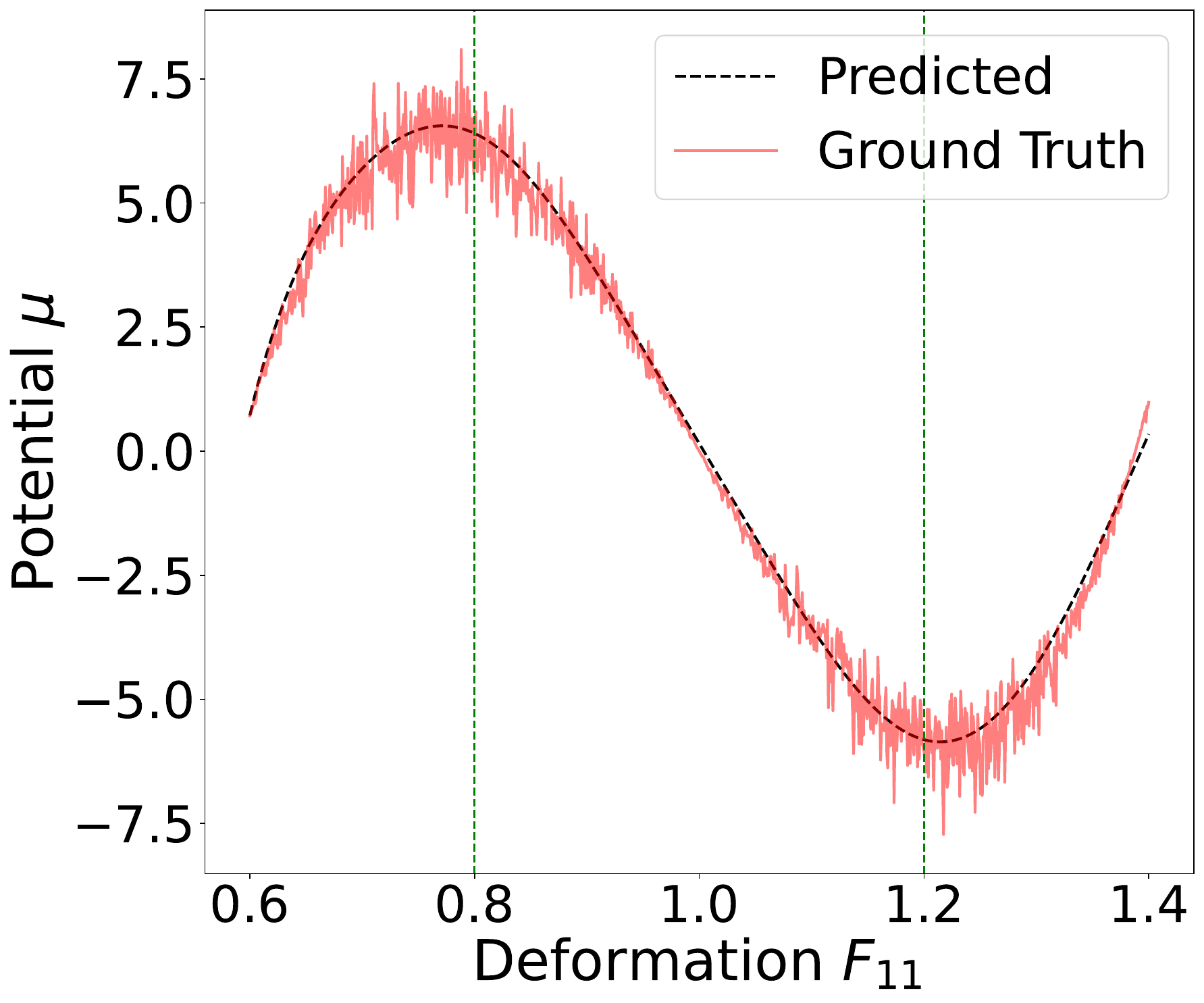}
\caption{Fits for $L_0$ (left), $L_1$ (center), and $L_2$ (right) regularizations.
Top row: free energy $\Psi$, middle row: stress $\Sb$, bottom row: chemical potential $\mu$.
Note only stress and chemical potential are in the training data.
Clearly the multiple minima of the potential are captured accurately.
}
\label{fig:mechchem_fits}
\end{figure}

\fref{fig:mechchem_fits} shows that MAP fits are comparably accurate for $L_0$, $L_1$, and $L_2$ regularization.
The models accurately predict the stress and chemical potential, as well as the free energy, which was not included in the training data. Here, the MAP model for the $L_0$ sparsification with $34$ parameters achieves accuracy comparable to that of the $ L_1$ and $ L_2$ regularized MAP models with $148$  parameters.
The $L_0$ sparsified expression is:
\begin{multline*}
\hat{\Psi} = 2.262 c - 1.734 e_{1} - 0.357 \log{\left(\frac{9.43 \left(\frac{e^{9.451 c}}{\left(1 + e^{- 0.506 c}\right)^{9.855}} + 1\right)^{1.085}}{\left(\frac{e^{18.73 c}}{\left(1 + e^{- 0.506 c}\right)^{7.063}} + 0.019\right)^{0.728}} + 1 \right)} \\
- 1.081 \log{\left(\frac{\left(\left(1 + e^{- 0.506 c}\right)^{5.615} e^{- 20.601 e_{2}} + 1\right)^{0.352} e^{0.02 c + 0.023 e_{6}}}{\left(\left(1 + e^{- 0.506 c}\right)^{4.612} e^{18.344 e_{2}} + 1\right)^{0.596}} + 1 \right)} \\
+ 0.296 \log{\left(9.766 \left(1 + e^{- 7.325 e_{1}}\right)^{4.219} \left(\frac{e^{18.73 c}}{\left(1 + e^{- 0.506 c}\right)^{7.063}} + 0.019\right)^{0.576} \right)} \\
\left(\left(1 + e^{- 0.506 c}\right)^{4.612} e^{18.344 e_{2}} + 1\right)^{5.65} \left(\left(1 + e^{- 0.506 c}\right)^{5.615} e^{- 20.601 e_{2}} + 1\right)^{6.124} \\ \left(e^{7.682 e_{1}} + 1\right)^{5.569} + 1
+ 0.485 \log{\left(\frac{1890.69}{\left(\frac{e^{9.451 c}}{\left(1 + e^{- 0.506 c}\right)^{9.855}} + 1\right)^{1.72}} + 1 \right)} - 15.98
\end{multline*}

\begin{figure}[H]
\centering
\includegraphics[width=0.44\textwidth]{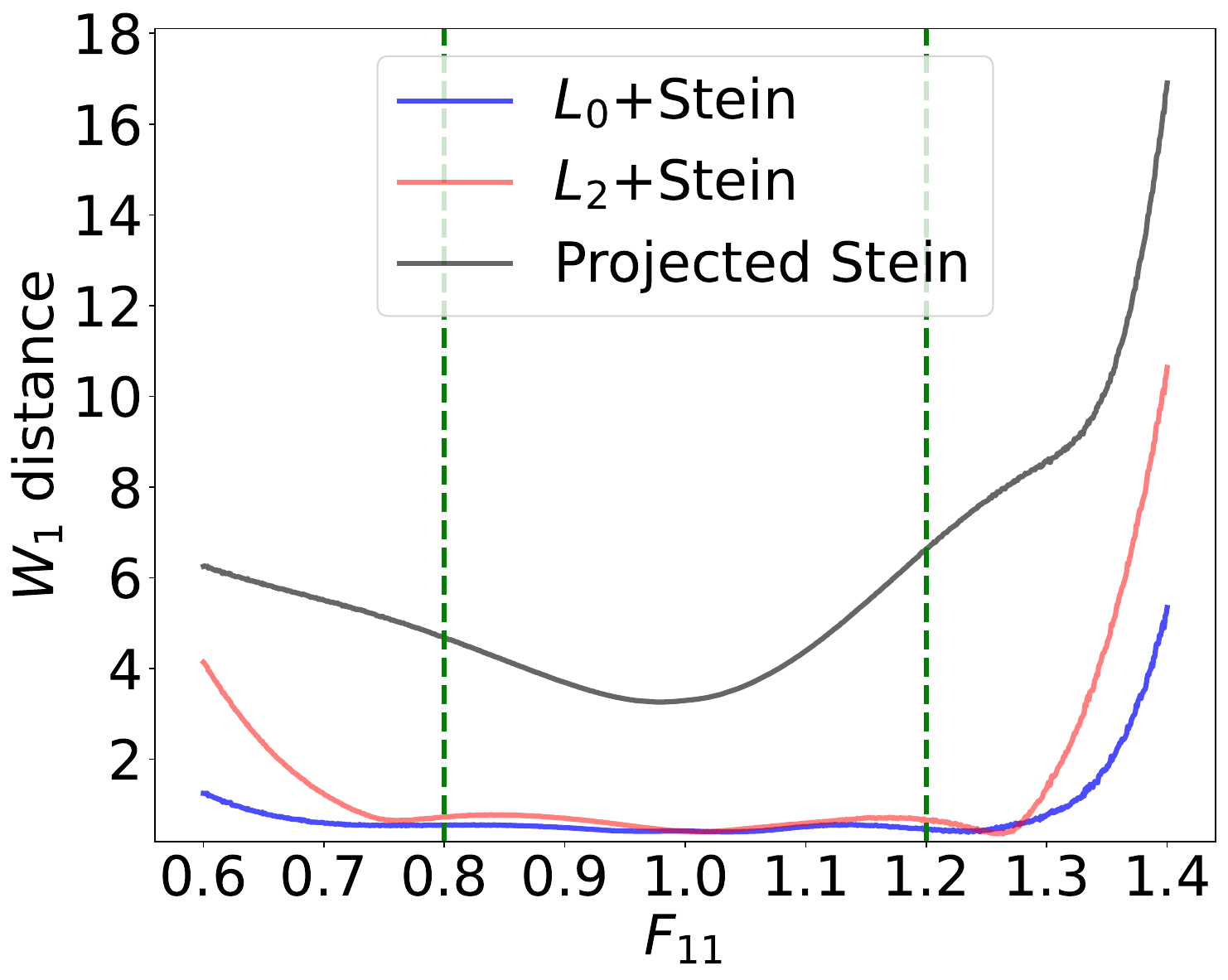}
\includegraphics[width=0.44\textwidth]{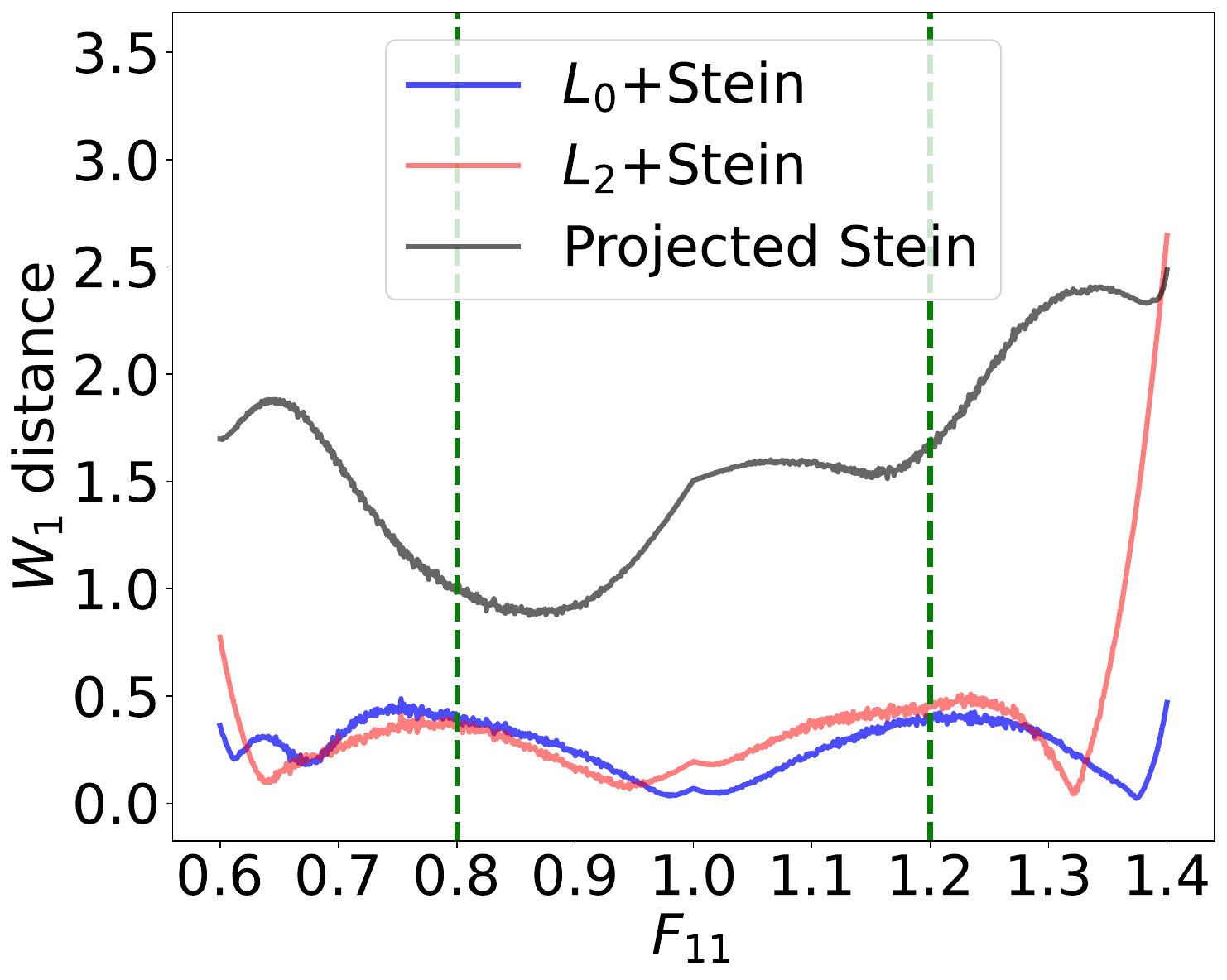}
\caption{Comparison of Wasserstein-1 distances for $L_0$ sparsified Stein, $L_2$ regularized Stein, and $L_2$ regularized projected Stein for stress (left) and chemical potential (right).}
\label{fig:mechchem_comparison}

\end{figure}

\fref{fig:mechchem_comparison} shows how the proposed method compares to the standard, regularized techniques.
The relative performance of the three methods is similar to that for the hyperelasticity demonstration, \fref{fig:hyperelasticity_sparsification}.
Clearly, the $L_0$+Stein approach is superior despite using only 50 particles versus 1000 for the other two methods.
Also, as shown in \tref{tab:mechchem_cost} the computational cost for the $L_0$+Stein approach is significantly lower than that of $L_2$+Stein approach.

\begin{table}[H]
\begin{tabular}{c|cc|cc|}
\cline{2-5}
& \multicolumn{2}{c|}{Hyperelasticity} & \multicolumn{2}{c|}{Mechanochemistry} \\ \hline
\multicolumn{1}{|c|}{Method} & \multicolumn{1}{c|}{Time {[}s{]}} & \begin{tabular}[c]{@{}c@{}}Number of \\ parameters \\ (Deterministic)\end{tabular} & \multicolumn{1}{c|}{Time {[}s{]}} & \begin{tabular}[c]{@{}c@{}}Number of \\ parameters \\ (Deterministic)\end{tabular} \\ \hline
\multicolumn{1}{|c|}{$L_0$+Stein} & \multicolumn{1}{c|}{1708} & 7 & \multicolumn{1}{c|}{14862} & 34 \\ \hline
\multicolumn{1}{|c|}{\begin{tabular}[c]{@{}c@{}}$L_2$+Projected \\ Stein\end{tabular}} & \multicolumn{1}{c|}{1794} & 102 & \multicolumn{1}{c|}{5127} & 148 \\ \hline
\multicolumn{1}{|c|}{$L_2$+Stein} & \multicolumn{1}{c|}{14816} & 1005 & \multicolumn{1}{c|}{165052} & 148 \\ \hline
\end{tabular}
\caption{Comparison of wall clock time and number of parameters for both the deterministic optimization and the Bayesian UQ for the hyperelasticity and mechanochemistry examples. All the models were trained on a single NVIDIA RTX A6000 GPU.}\label{tab:mechchem_cost}
\end{table}

\section{Conclusion} \label{sec:conclusion}

For highly parameterized models like NNs, we proposed an alternative SVGD method to pSVGD that embeds more aspects of the posterior parameter manifold than linearization can provide.
We demonstrated the advantages of the method on applications from mechanics.
For one example we exploited the constraint of polyconvexity of the underlying potential, while the other was distinctly non-convex.
For these examples, $L_0$ sparsification of the NN model prior to applying SVGD demonstrated superior performance to alternative regularizations and model reduction techniques.

\begin{figure}
\centering
\includegraphics[width=0.3\textwidth]{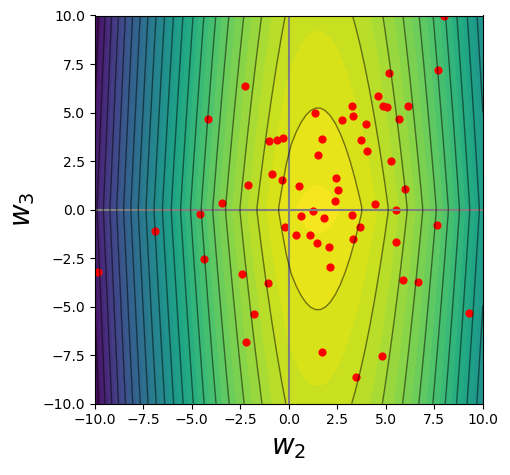}
\includegraphics[width=0.3\textwidth]{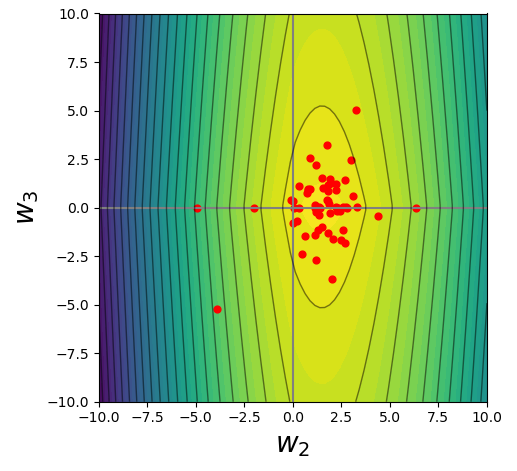}
\includegraphics[width=0.3\textwidth]{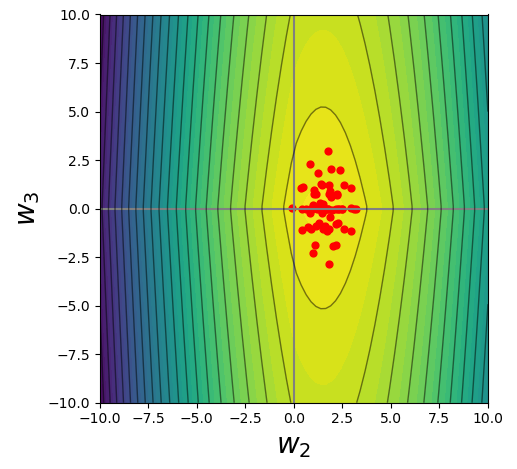}
\caption{Demonstration of gradient descent with sparsifying prior, epoch=0 (left), 500 (middle), 1000 (right).
$L_{1}$ prior and multivariate likelihood with precision
$
\begin{bmatrix}
2 & 1 & 0 \\
1 & 2 & 0 \\
0 & 0 & 0.025 \\
\end{bmatrix}
$ and mean $
\begin{bmatrix}
1 & 2 & 3
\end{bmatrix}^T
$.}
\label{fig:L1_GD_demo}
\end{figure}

In future work, we want to use the uncertainty information from the proposed $L_0$ Stein technique in forward propagation studies of large-scale finite element simulations \cite{jones2021minimally,de2023accurate}.
Since each Stein particle represents a model realization, this should be straightforward.
In addition, we will pursue concurrent UQ and model sparsification by augmenting the Stein gradient \eref{eq:stein_gradient} with the gradient of a sparsifying prior.
\fref{fig:L1_GD_demo} demonstrates the convergence of an ensemble of particles in this augmented gradient flow with a $L_1$ prior.
Clearly, the particles cluster around the data mean where the likelihood has high precision and are forced to zero where the likelihood precision in that parameter is low.
This approach may have computational advantages when it is not feasible to find a sparse model first.
Lastly, since $L_0$+Stein readily provides uncertainty information, it can be used in active learning based on UQ objectives such as the upper confidence bound and expected information gain \cite{li2006confidence,settles2011theories}, which we wish to exploit in practical applications.

\section*{Acknowledgements}
The authors wish to acknowledge Prof. Krishna Garikipati (USC) for pointing us to the numerical illustration of mechanochemistry.
GAP and NB were supported by the SciAI Center, and funded by the Office of Naval Research (ONR), under Grant Number N00014-23-1-2729.
REJ and CS were supported by the U.S. Department of Energy, Advanced Scientific Computing program.
CS was also supported by the Scientific Discovery through Advanced Computing (SciDAC) program through the FASTMath Institute.
Sandia National Laboratories is a multi-mission laboratory managed and operated by National Technology \& Engineering Solutions of Sandia, LLC (NTESS), a wholly owned subsidiary of Honeywell International Inc., for the U.S. Department of Energy’s National Nuclear Security Administration (DOE/NNSA) under contract DE-NA0003525. This written work is authored by an employee of NTESS. The employee, not NTESS, owns the right, title and interest in and to the written work and is responsible for its contents. Any subjective views or opinions that might be expressed in the written work do not necessarily represent the views of the U.S. Government. The publisher acknowledges that the U.S. Government retains a non-exclusive, paid-up, irrevocable, world-wide license to publish or reproduce the published form of this written work or allow others to do so, for U.S. Government purposes. The DOE will provide public access to results of federally sponsored research in accordance with the DOE Public Access Plan.



\appendix
\section{Stein Variational Gradient Descent}\label{app:SVGD}

Given a set of data $\mathcal{D} =\{\inputvector^{i},\outputvector^{i}\},_{i=1}^{N}$, with the likelihood function $\prob(\outputvector|\inputvector,\parameters)$ and prior $\prob(\parameters)$, as in \sref{sec:bayes}, we are interested in obtaining the posterior distribution $\prob(\parameters|\mathcal{D})$.
SVGD \cite{liu2016stein} aims to approximate the posterior distribution with a variational distribution $q^*(\parameters)$, which lies in the restricted set of distributions $\mathcal{Q}$:
\begin{equation} \label{eq:surrogate_dist2}
q^{*}(\parameters)
=\underset{q \in \mathcal{Q}}{\arg \min }\, {\KLdiv}(q(\parameters) \| \prob(\parameters | \mathcal{D}))
=\underset{q \in \mathcal{Q}}{\arg \min}\, \mathbb{E}_{q}[\log q(\parameters)-\log \tilde{p}(\parameters | \mathcal{D})] \ ,
\end{equation}
where $\tilde{p}(\parameters | \mathcal{D})= \prob(\mathcal{D} | \parameters) \prob(\parameters)=\prod_{i=1}^{N} \prob\left(\mathbf{y}^{i} | \parameters, \mathbf{x}^{i}\right) \prob(\parameters)$ is the unnormalized posterior.
The normalization constant associated with the evidence is not considered when we optimize the Kullback-Liebler (KL) divergence defined in \eref{eq:surrogate_dist2}.
In SVGD, we take an initial tractable distribution represented in terms of samples and then apply a transformation to each of these samples:
\begin{equation}
\bm{T}(\parameters)=\parameters+\epsilon \bm{\phi}(\parameters),
\end{equation}
where $\epsilon$ is the step size and $\bm{\phi}(\parameters) \in \mathcal{F}$ is the perturbation direction within a function space $\mathcal{F}$.
Therefore, $\bm{T}$ transforms the initial density $q(\parameters)$ to $q_{[\bm{T}]}(\parameters)$
\begin{equation}
q_{[\bm{T}]}(\parameters)=q\left(\bm{T}^{-1}(\parameters)\right)\left|\operatorname{det}\left(\nabla \bm{T}^{-1}(\parameters)\right)\right|.
\end{equation}

Unlike the more common mean field variational inference, SVGD uses a particle approximation for the variational posterior rather than a parametric form.
Therefore, we consider a set of samples $\{\parameters_i\}_{i=1}^{S}$,  with the empirical measure:
\begin{equation}
\mu_{S}(d\parameters) = \frac{1}{S}\sum_{i=1}^{S}\delta(\parameters-\parameters^{i})d\parameters \ .
\end{equation}
where $S$ is the total number of samples.
While the empirical measure $\mu_S$ converges weakly to the true measure $\mu$ as the number of samples $S$ increases, it is important for the measure $\mu$ to weakly converge to the measure $\nu_\prob(\mathrm{d} \parameters) = \prob(\parameters \giv \data)\mathrm{d}\parameters$ of the true posterior $p \equiv  \prob(\parameters \giv \data)$.
The minimum KL divergence of the variational approximation and the target distribution:
\begin{equation}
\min_{\bm{\phi} \in \mathcal{F}}\Big[ \frac{\mathrm{d}}{\mathrm{d} \epsilon} \KLdiv (\bm{T}\mu||\nu_{\prob})|_{\epsilon=0} \Big].
\end{equation}
under the transformation $\mu \to \bm{T}\mu$, determines the optimal $\phib$ and approximate posterior.
This term can also be expressed as~\cite{liu2016stein}:
\begin{equation}
\frac{\mathrm{d}}{\mathrm{d} \epsilon} \KLdiv (\bm{T}\mu||\nu_{\prob})|_{\epsilon=0} = -\mathbb{E}_{\mu}[\mathcal{T}_\prob\bm{\phi}],
\end{equation}
where $\mathcal{T}_{\prob}$ is the Stein operator associated with the distribution $\prob$  given by:
\begin{equation}
\mathcal{T}_{\prob} \bm{\phi} = \frac{\nabla \cdot (\prob\bm{\phi})}{\prob} = \frac{(\nabla \prob)\cdot \bm{\phi}+\prob(\nabla\cdot \bm{\phi})}{\prob} = (\nabla \log \prob)\cdot \bm{\phi} + \nabla \cdot \bm{\phi}.
\end{equation}
The term $\mathbb{E}_{\mu}\left[\mathcal{T}_{\prob} \bm{\phi}\right]$ evaluates the difference between the measures $\nu_\prob$ and $\mu$ and its maximum is defined as the Stein discrepancy $(\mathcal{S}(\mu, \prob))$:
\begin{equation}
\mathcal{S}(\mu, \prob)=\max _{\bm{\phi} \in \mathscr{F}} \mathbb{E}_{\mu}\left[\mathcal{T}_{\prob} \bm{\phi}\right].
\end{equation}

If the functional space $\mathcal{F}$ is chosen to be the unit ball in a product reproducing kernel Hilbert space with the positive kernel $\kernel(\parameters,\parameters^{'})$, then the  Stein discrepancy has a closed-form solution~\cite{liu2016stein}:
\begin{equation} \label{eq:stein_gradient2}
\bm{\phi}^{*}(\parameters)
\propto \mathbb{E}_{\parameters^{'}\sim \mu}[\mathcal{T}^{\parameters^{'}}_{\prob}\kernel(\parameters,\parameters^{'})]
= \mathbb{E}_{\parameters^{'}\sim \mu}[\nabla_{\parameters^{'}}\log \prob(\parameters^{'} \giv \mathcal{D})\kernel(\parameters,\parameters^{'})+\nabla_{\parameters^{'}}\kernel(\parameters,\parameters^{'})],
\end{equation}

\section{Input convex neural network} \label{app:icnn}

In \cref{amos2017input}, an ICNN network is defined as follows: for an output $\outputvector$ an corresponding input $\inputvector$, the  neural network $\mathcal{N}$ with $N$ number of layers is simply:
\begin{eqnarray}
\hiddenvector_1 &=& \sigma_1 \left(
\Ws_1 \inputvector +\bs_1
\right) \nonumber \\
\hiddenvector_k &=& \sigma_k \left(
\Vs_k \inputvector + \Ws_k \hiddenvector_{k-1} +\bs_k
\right)  \qquad k=2, \ldots, N-1  \label{eq:ICNN} \\
\outputvector &=& \Vs_N \inputvector + \Ws_N \hiddenvector_{N-1} +\bs_N   \nonumber
\end{eqnarray}
with weights $\Ws_k$ and $\Vs_k$, activation functions $\sigma_k$ and $1<k<N$.
The weights and biases form the set of trainable parameters $\parameters = \lbrace \Ws_k,  \Vs_k, \bs_k \rbrace$.
The output is convex with respect to the input if the weights $\Ws_k$ are non-negative and the activation functions $\sigma_k$ are convex and non-decreasing \cite{amos2017input}.

In this work, to be computationally efficient and to obtain faster convergence, we ignore bias terms and assign the same weight $\Vs_k=\Ws_k$ per layer for  $\Vs_k$ while having the weight values $\Ws_k$ to be non-negative.
Here, we consider \emph{softplus}  as the activation functions $\sigma_k$:
\begin{eqnarray}
\hiddenvector_1 &=& \sigma_1 \left(
\Ws_1 \inputvector
\right) \nonumber \\
\hiddenvector_k &=& \sigma_k \left(
\Ws_k \inputvector + \Ws_k \hiddenvector_{k-1}
\right) = \sigma_k \left(
\Ws_k (\inputvector + \hiddenvector_{k-1})
\right) \hspace{2mm} k=2, \ldots, N-1  \label{eq:ICNN1} \\
\outputvector &=& \Ws_N \inputvector + \Ws_N \hiddenvector_{N-1} = \Ws_N (\inputvector + \hiddenvector_{N-1}) ,  \nonumber
\end{eqnarray}

\end{document}